\documentclass[10pt,twocolumn,letterpaper]{article}

\usepackage[pagenumbers]{cvpr} 

\definecolor{cvprblue}{rgb}{0.21,0.49,0.74}
\usepackage[pagebackref,breaklinks,colorlinks,allcolors=cvprblue]{hyperref}

\def\paperID{1} 
\def\confName{CVPR}
\def\confYear{2025}

\usepackage[utf8]{inputenc} 
\usepackage[T1]{fontenc}    
\usepackage{hyperref}       
\usepackage{url}            
\usepackage{booktabs}       
\usepackage{amsfonts}       
\usepackage{nicefrac}       
\usepackage{microtype}      
\usepackage{xcolor}         
\usepackage{amsmath}
\usepackage{multirow}
\usepackage{fontawesome}
\usepackage{lipsum}

\usepackage{colortbl}
\usepackage{times}
\usepackage{graphicx}
\usepackage{amsmath}
\usepackage{amssymb}
\usepackage{hyperref}
\usepackage{setspace} 

\usepackage{xfakebold}

\newcommand{\fbseries}{\unskip\setBold\aftergroup\unsetBold\aftergroup\ignorespaces}
\makeatletter
\newcommand{\setBoldness}[1]{\def\fake@bold{#1}}
\makeatother

\usepackage{enumitem} 
\usepackage[accsupp]{axessibility}  

\usepackage[nolist,nohyperlinks]{acronym}

\begin{acronym}
\acro{df24}[DF24]{Danish Fungi 2024}
\acro{df20}[DF20]{Danish Fungi 2020}
\acro{df24m}[DF24M]{Danish Fungi 2024 - Mini}
\acro{df20m}[DF20M]{Danish Fungi 2020 - Mini}
\acro{df24f}[DF24F]{Danish Fungi 2024 - Few-Shot}

\acro{fungi}[FungiTastic]{FungiTastic}
\acro{fungim}[FungiTastic--M]{FungiTastic--Mini}
\acro{fungif}[FungiTastic--FS]{FungiTastic--Few-shot}
\acro{fungios}[FungiTastic--OS]{FungiTastic--Open-set}

\acro{sam}[SAM]{Segment Anything Model}
\acro{ce}[CE]{Cross-Entropy}
\acro{iou}[IoU]{Intersection over Union}
\acro{vit}[ViT]{Vision Transformer}
\acro{bce}[BCE]{Binary Cross Entropy}
\acro{fgvc}[FGVC]{Fine-Grained Visual 
Classification}
\acro{tta}[TTA]{Test-Time Adaptation}
\end{acronym}

\title{FungiTastic: A Multi-Modal Dataset and Benchmark for Image Categorization}

\author{%
  \textbf{ Lukas Picek \hspace{-2.5pt}\raisebox{3pt}{\mbox{{\includegraphics[width=3.5mm, height=3.5mm]{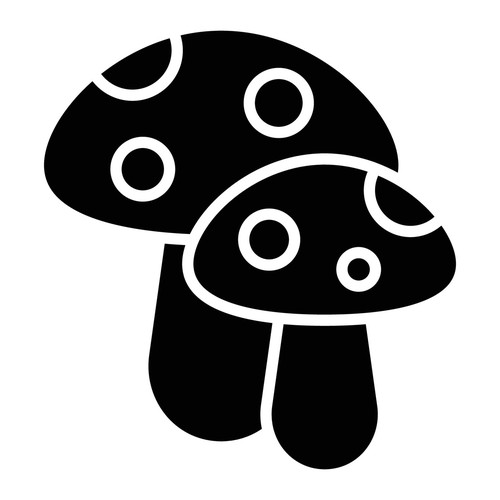}}}}, 
  Kl\'{a}ra Janou\v{s}kov\'{a} \hspace{-2.5pt}\raisebox{3pt}{\mbox{{\includegraphics[width=3.5mm, height=3.5mm]{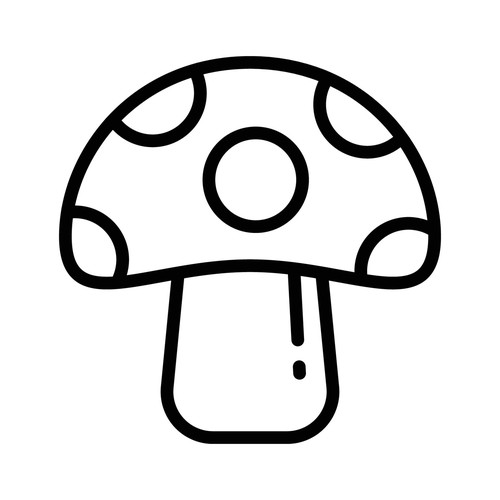}}}},
  Vojtech Cermak \hspace{-2.5pt}\raisebox{3pt}{\mbox{{\includegraphics[width=3.5mm, height=3.5mm]{images/mushroom3.jpg}}}}, 
  and Jiri Matas \hspace{-2.5pt}\raisebox{3pt}{\mbox{{\includegraphics[width=3.5mm, height=3.5mm]{images/mushroom3.jpg}}}}} \\
  \hspace{-2.5pt}\raisebox{3pt}{\mbox{{\includegraphics[width=2.5mm, height=2.5mm]{images/mushrooms-b.jpg}}}} \hspace{-2.5pt}University of West Bohemia \& Inria,
  \hspace{-2.5pt}\raisebox{3pt}{\mbox{{\includegraphics[width=2.5mm, height=2.5mm]{images/mushroom3.jpg}}}}\hspace{-2.5pt} CTU in Prague\\ 
  {\small\texttt{lukaspicek@gmail.com, \{janoukl1,cermavo3,matas\}@fel.cvut.cz}} 
}

\begin{document}


\twocolumn[{%
\renewcommand\twocolumn[1][]{#1}%
\maketitle
\begin{center}
\vspace{-0.25cm}

    \includegraphics[width=\linewidth]{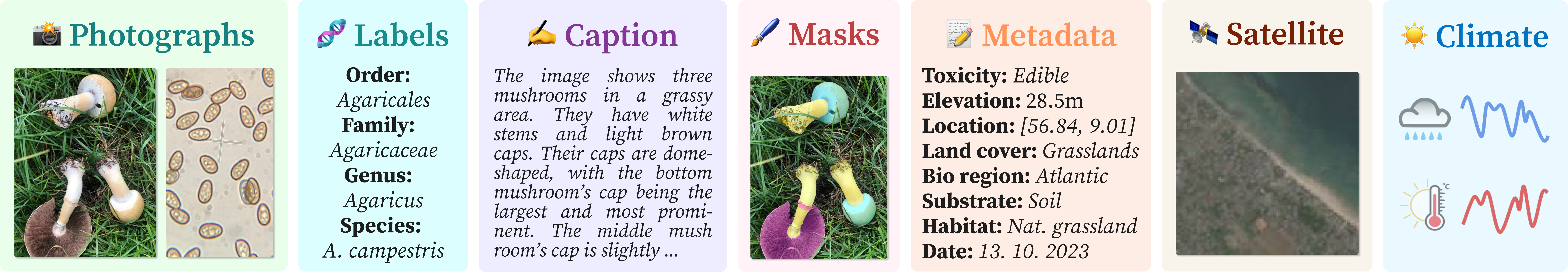}

    \captionof{figure}{\textbf{A \acl{fungi} observation} includes 
        {\setlength{\fboxsep}{1pt}\colorbox{green!10}{one or more photos}} of an \textit{observed} specimen with expert-verified taxon labels (some DNA sequenced) and occasionally also a microscopic image of its spores. 
        {\setlength{\fboxsep}{1pt}\colorbox{blue!7}{Textual captions}},   
        {\setlength{\fboxsep}{1pt}\colorbox{Apricot!20}{observation metadata}}, {\setlength{\fboxsep}{1pt}\colorbox{brown!10}{geospatial data}}, and {\setlength{\fboxsep}{1pt}\colorbox{cyan!7}{climatic time-series data}} are available for virtually
        all observations. For a subset ($\sim$70k photos), we provide 
        {\setlength{\fboxsep}{1pt}\colorbox{magenta!7}{body part segmentation masks}}.
    }
    \label{fig:observation}
    \vspace{0.75cm}
\end{center}
}]

\begin{abstract}

We introduce a new, challenging benchmark and a dataset, FungiTastic, based on fungal records
continuously collected over a twenty-year span. The dataset is labelled and curated by experts and consists of about 350k multimodal observations 
of 6k fine-grained categories (species). The fungi observations include photographs and additional data, e.g., meteorological and climatic data, satellite images, and body part segmentation masks.
FungiTastic is one of the few benchmarks that include a test set with DNA-sequenced ground truth of unprecedented label reliability.
The benchmark is designed to support 
(i) standard closed-set classification, 
(ii) open-set classification,
(iii) multi-modal classification,
(iv) few-shot learning, 
(v) domain shift, and many more.
We provide tailored baselines for many use cases,
a multitude of ready-to-use pre-trained models on 
\href{https://huggingface.co/collections/BVRA/fungitastic-66a227ce0520be533dc6403b}{HuggingFace},
and a framework for model training.
The documentation and the baselines are available at 
\href{https://github.com/BohemianVRA/FungiTastic/}{GitHub} 
and \href{https://www.kaggle.com/datasets/picekl/fungitastic}{Kaggle}.
\end{abstract}

\section{Introduction}
\label{sec:intro}
Biological problems provide a natural, challenging setting for benchmarking image classification methods \cite{picek2022danish,van2021benchmarking,van2018inaturalist,sastry2025taxabind}.
Consider the following aspects inherently present in biological data. The species distribution is typically seasonal and constantly evolving under the influence of external factors such as precipitation levels, temperature, and loss of habitat, exhibiting constant {\it domain shifts}.
Species categorization is fine-grained, with high intra-class and low inter-class variance. 
The distribution is often long-tailed; only a few samples are available for rare species (\textit{few-shot learning}). 
New species are being discovered, raising the need for the ``\textit{unknown}'' class option (i.e., \textit{open-set recognition}).
Commonly, the set of classes has a hierarchical structure, and different misclassifications may have very different costs (\textit{i.e., non-standard losses}).
Think of mistaking a poisonous mushroom for an edible, potentially lethal, and an edible mushroom for a poisonous one, which at worst means an empty basket. 
Similarly, needlessly administering anti-venom after making a wrong decision about a harmless snake bite may be unpleasant, but its consequences are incomparable to not acting after a venomous bite. 

Common benchmarks \cite{deng2009imagenet,wah2011caltech,maji2013fine,van2018inaturalist} generate independent and identically distributed (i.i.d.)  data by shuffling and randomly splitting it for training and evaluation.  In real-world applications, i.i.d data are rare since training data are collected well before deployment and everything changes over time \cite{wiki:Heraclitus}. Moreover, they fail to address  the above-mentioned aspects important in many instance of ML system deployment:
robustness to distribution and domain shifts,  ability to detect classes not represented in the training set,  limited training data, and dealing with non-standard losses.

For benchmarking, it is crucial to ensure that methods are tested on data not indirectly ``seen'', without knowing\,\,\cite{goodfellow2016deep,hendrycks2019benchmarking}, especially given the huge dataset used for training LLMs or VLMs and possibly covering the entirety of the internet at a certain point in time.
Conveniently, many domains in nature are of interest to experts and the general public, who provide a continuous stream of new and annotated data \cite{picek2022automatic,swanson2015snapshot}.
The public's involvement introduces the problem of noisy training data; evaluating the robustness of this phenomenon is also of practical importance.

In the paper, we introduce \textbf{\acl{fungi}}, a multi-modal dataset and benchmark based on fungi observations\footnote{A set of photographs and additional metadata describing one particular fungi specimen and surrounding environment. Usually, each photograph focuses on a different organ. For an example observation, see Figure \ref{fig:observation}}, which takes  
advantage of the favorable properties of natural data discussed above and shown in Figure \ref{fig:observation}.
The fungi observations include photographs, satellite images, meteorological data, segmentation masks, textual captions, and location-related metadata.
The \textit{location metadata} enriches the observations with attributes such as the timestamp, GPS location, and information about the substrate and habitat.

By incorporating various modalities, the dataset supports a robust benchmark for multi-modal classification, enabling the development and evaluation of sophisticated machine-learning models under realistic and dynamic conditions.

\noindent
\textbf{The key contributions of the \acl{fungi} benchmark are:}
\begin{itemize}
[left=4pt]
\vspace{1mm}
    \item It addresses real-world challenges such as domain shifts, open-set problems, and few-shot classification, providing a realistic benchmark for developing robust ML models. \vspace{2mm}

    \item The proposed benchmarks allow for addressing fundamental problems beyond standard image classification, such as novel-class discovery, few-shot classification, and 
    evaluation with non-standard cost functions.   \vspace{2mm}

    \item It includes diverse data types, such as photographs, satellite images, bioclimatic time-series data, segmentation masks, contextual metadata (e.g., timestamp, camera metadata, location, substrate, and habitat), and image captions, offering a rich, multimodal benchmark.

\end{itemize}

\section{Related Work}
Classification of data originating in nature, including 
images of birds \cite{berg2014birdsnap,wah2011caltech}, plants \cite{garcin2021pl,goeau2017plant}, snakes \cite{bolon2022artificial,picek2022overview}, fungi \cite{picek2022danish,van2018inaturalist}, and insects \cite{gharaee2024step,nguyen2024insect}
has been widely used to benchmark machine learning algorithms, not just fine-grained visual categorization. The datasets were instrumental in focusing on fine-grained recognition and attracting attention to challenging natural problems.

However, the datasets are typically artificially sampled, solely image-based, and focused on traditional image classification. 
Most commonly used datasets are small by modern standards, with a limited number of categories, which restricts their usefulness for large-scale and highly diverse applications. Though performance being often saturated, reaching an accuracy of 85--95 \% (rightmost column of Tab.~\ref{tab:fgvc_data_comp}), these datasets are still widely used in the community and have reached thousands of citations in the past few years.
Many popular datasets also suffer from specific limitations that compromise their generalizability and robustness. Common issues include:
\begin{itemize}
[left=4pt]
\vspace{1mm}
    \item \textbf{Lack of Multi-Modal Data}: Available datasets are predominantly image-based, with few offering auxiliary metadata like geographic or temporal context, which is essential for real-world applications where distribution changes and context is important. \vspace{1mm}

    \item \textbf{Biases in Data Representation}: Many datasets exhibit regional and other biases \cite{stock2018convnets}, which can lead to biased models that do not perform well across different populations or environments. This lack of diversity can severely limit the usability of models trained on these datasets for global applications. \vspace{1mm}
    
    \item \textbf{Single task focus}: While current ML applications require adaptability to tasks such as open-set classification, few-shot learning, and out-of-distribution detection, many of these datasets were not designed with these tasks in mind, limiting their usefulness for modern benchmarking. \vspace{1mm}

    \item \textbf{Labeling Errors and Quality Control}: Label errors are prevalent in widely-used datasets \cite{beyer2020we, van2015building}. Mislabeling, especially in fine-grained categories, can reduce the reliability of these datasets as benchmarks and reduce the model’s ability to learn fine distinctions.
\end{itemize}

\begin{table}[!b]
\setlength{\tabcolsep}{2.5pt}
\footnotesize
    \centering
        \caption{\textbf{\textit{Resent} and popular fine-grained classification datasets}. We list suitability for closed-set ({C}),
        open-set ({OS}),
        and few-shot ({F}) classification,
        segmentation ({S}),
        out-of-distribution ({OOD}) and
        multi-modal ({M$^2$}) evaluation. Modalities, e.g., images (I), metadata (M), and masks (S), are available for training.
        The SOTA accuracy is limited to the classification task. For TaxaBench-8k, we report zero-shot performance. 
        {$\forall =  \text{ \{C, OS, FS, S, OOD, M$^2$\}}$}}
        \vspace{-0.1cm}
\begin{tabular}{@{}l@{\hspace{2pt}}rr@{\hspace{8pt}}c@{\hspace{4pt}}c@{\hspace{4pt}}ccl@{}}
\toprule
   &  & & \multicolumn{3}{c}{\textbf{Modals.}} &  & \textbf{SOTA}$^\dagger$ \\
\textbf{Dataset} & \textbf{Classes} & \textbf{Images} & I & M & S & \textbf{Tasks} & Accuracy \\
\midrule
Oxford-Pets \cite{parkhi2012cats}  & 37 & ~5k & \checkmark & -- & -- & {\scriptsize C} & 97.1\,\cite{foret2020sharpness} \\
FGVC Aircraft \cite{maji2013fine} & 102 & ~10k & \checkmark & -- & -- & {\scriptsize C} & 95.4  \cite{bera2022sr} \\
Stanford Dogs \cite{khosla2011novel} & 120 & ~20k & \checkmark & -- & -- & {\scriptsize C} & 97.3  \cite{bera2022sr} \\
Stanford Cars \cite{krause20133d} & 196 & ~16k & \checkmark & -- & -- & {\scriptsize C} & 97.1  \cite{liu2023learn} \\
Species196 \cite{he2024species196}  & 196 &  20k & \checkmark & \checkmark & -- & {\scriptsize C$/$M$^2$} & 88.7 \cite{he2024species196} \\
CUB-200-2011 \cite{wah2011caltech}  & 200 & ~12k & \checkmark & \checkmark & \checkmark & {\scriptsize C}  & 93.1  \cite{chou2023fine} \\
NABirds \cite{van2015building}  & 555 & 49k & \checkmark & -- & -- & {\scriptsize C$/$F$/$M$^2$} & 93.0  \cite{diao2022metaformer} \\
PlantNet300k \cite{garcin2021pl}  & 1,081 & ~275k & \checkmark & -- & -- & {\scriptsize C} & 92.4  \cite{garcin2021pl} \\
DanishFungi2020 \cite{picek2022danish}  & 1,604 & 296k & \checkmark & \checkmark & -- & {\scriptsize ~C$/$M$^2$} & 80.5  \cite{picek2022danish} \\
ImageNet-1k \cite{deng2009imagenet}  & 1,000 & 1.4m & \checkmark & -- & -- & {\scriptsize C$/$FS} & 92.4 \cite{dong2023peco} \\
TaxaBench-8k \cite{sastry2025taxabind}  & 2225 & 9k & \checkmark & \checkmark & -- & {\scriptsize ~C$/$M$^2$} & 37.5\, \cite{sastry2025taxabind} \\
iNaturalist \cite{van2018inaturalist}  & 5,089 & 675k & \checkmark & -- & -- & {\scriptsize C$/$FS}  & 93.8  \cite{srivastava2024omnivec} \\
ImageNet-21k \cite{ridnik2021imagenet}  & 21,841 & 14m & \checkmark & -- & -- & {\scriptsize C$/$FS} & 88.3  \cite{srivastava2024omnivec} \\
Insect-1M \cite{nguyen2024insect}  & 34,212 & 1m & \checkmark & \checkmark & -- & {\scriptsize ~C$/$M$^2$} & ~~\,-- \\
\midrule
(our) \acl{fungi} & 2,829 & 620k & \checkmark & \checkmark & \checkmark & $\forall$& 75.3 \\
(our) \acl{fungim} &  215 & 68k & \checkmark & \checkmark & \checkmark &  $\forall$& 74.8 \\
\bottomrule
\end{tabular}
    \label{tab:fgvc_data_comp}

\end{table}

\newpage
\section{The FungiTastic Benchmark}
\label{sec:dataset_desr}

\acs{fungi} is built from fungi observations submitted to the Atlas of Danish Fungi before the end of 2023, which were labeled by taxon experts on a species level. In total, more than 350k observations consisting of 630k photographs collected over 20 years are used. Apart from the photographs, each observation includes additional observation data (see Figure \ref{fig:observation}) ranging from satellite images, meteorological data, and tabular metadata (e.g., timestamp, GPS location, and information about the substrate and habitat) to segmentation masks and toxicity status. The vast majority of observations got all of the attributes annotated. For details about the attribute description and its acquisition process, see Subsection \ref{subsec:obs_data}. 
Since the data comes from a long-term conservation project, its seasonality and naturally shifting distribution make it suitable for time-based splitting. 
In this so-called \textbf{temporal division}, all data collected up to the end of 2021 is used for \textbf{training}, while data from 2022 and 2023 is reserved for \textbf{validation} and \textbf{testing}, respectively.

The FungiTastic benchmark is designed to go beyond standard closed-set classification and support a wide range of challenging machine learning tasks, including (i) open-set classification, (ii) few-shot learning, (iii) multi-modal learning, and (iv) domain shift evaluation. To facilitate these tasks, we provide several curated subsets, each tailored for specific experimental setups. A general overview of these subsets is provided below, with detailed statistics and further information in the Appendix (see Table~\ref{tab:FungiTastic_subsets}). \\

\noindent\textbf{\acs{fungi}} is a general subset that includes around 346k observations of 4,507 species accompanied by a wide set of additional observation data. The FungiTastic has dedicated validation and test sets specifically designed for closed-set and open-set scenarios. While the closed-set validation and test sets only include species present in the training set, the open-set also includes observations with species observed only after 2022 (validation) and 2023 (test), i.e., species not available in the training set. All the species with no examples in the training set are labeled as "\textit{unknown}". \textbf{Additionally, we include a DNA-based test set} of 725 species and 2,041 observations.  \\

\noindent\textbf{\ac{fungim}} is a compact and challenging subset of the FungiTastic dataset designed primarily for prototyping and consisting of all observations belonging to 6 hand-picked genera (e.g., \textit{Russula}, \textit{Boletus}, \textit{Amanita}, \textit{Clitocybe}, \textit{Agaricus}, and \textit{Mycena})\footnote{These genera produce fruiting bodies of the toadstool type, which include many visually similar species and are of significant interest to humans due to their common use in gastronomy.}. This subset comprises 67,848 images (36,287 observations) of 253 species, greatly reducing the computational requirements for training. Exclusively, we include body part mask annotations.

\noindent\textbf{\acs{fungif}} subset, FS for few-shot, is formed by species with less than 5 observations in the training set, which were removed from the main (\acs{fungi}) dataset. The subset contains 6,391 observations encompassing 12,015 images of a total of 2,427 species. As in the \acs{fungi} data, the split into validation and testing is done according to the year of acquisition. 

\subsection{Additional Observation Data}
\label{subsec:obs_data}
This section provides an overview of the accompanying data available for virtually all user-submitted observations. 
For each type, we describe the data itself and, if needed, its acquisition process as well.
Below, we describe 
\textbf{(i) tabular metadata}, which includes key environmental attributes and taxonomic information for nearly all observations,
\textbf{(ii) remote sensing data} at fine-resolution geospatial scale for each observation site,
\textbf{(iii) meteorological data}, which provides long-term climate variables, 
\textbf{(iv) body part segmentation masks} that delineate specific morphological features of fungi fruiting bodies, such as caps, gills, pores, rings, and stems, and
\textbf{(v) image captions}.
All that metadata is integral to advancing research combining visual, textual, environmental, and taxonomic information. \\

\begin{figure}[h]
\vspace{-0.1cm}
    \centering
    \includegraphics[width=0.95\linewidth]{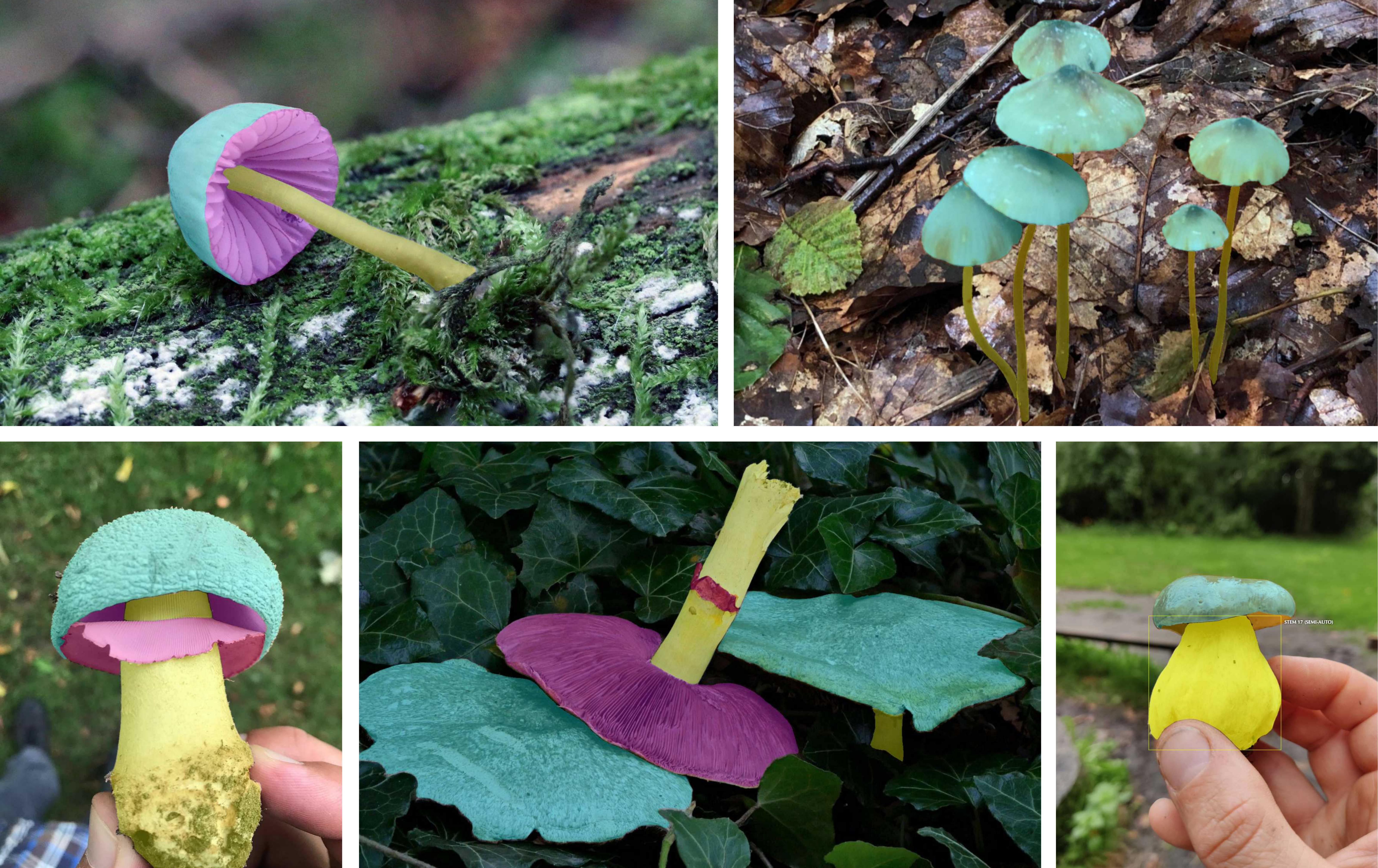}
    \caption{\textbf{FungiTastic body part segmentation}. We consider five different categories, e.g., the cap, gills, stem, pores, and the ring.}
    \label{fig:segments}
    \vspace{-0.1cm}
\end{figure}

\noindent\textbf{Body part segmentation masks} of fungi fruiting bodies are essential for accurate identification and classification \cite{deacon2013fungal}. These morphological features provide crucial taxonomic information distinguishing some visually similar species. 
Therefore, we provide instance segmentation masks for all photographs in the \ac{fungim}. We consider various semantic categories such as \textit{cap}, \textit{gills}, \textit{pores}, \textit{ring}, \textit{stem}, etc. These annotations (see Figure \ref{fig:segments}) are expected to drive advances in interpretable recognition methods \cite{rigotti2021attention} and evaluation \cite{Hesse_2023_ICCV}, with masks also enabling instance segmentation for separate foreground and background modeling \cite{bhatt2024mitigating}.
All segmentation mask annotations were semi-automatically generated in \href{https://github.com/cvat-ai/cvat}{CVAT} using the Segment Anything Model \cite{Kirillov_2023_ICCV} and human supervision, i.e., annotators fixed all wrong masks.

\noindent\textbf{Multi-band remote sensing data} offer detailed and globally consistent environmental information at a fine resolution, making it a valuable resource for species categorization (i.e., identification) \cite{rocchini2016satellite} and species distribution modeling \cite{botella2018deep,picek2024geoplant}. To allow testing the potential of such data and to facilitate easy use of geospatial data, we provide multi-band (e.g., R, G, B, NIR, elevation, and landcover) satellite patches with 64$\times$64 pixel resolution at 10m spatial resolution per pixel for (elevation and landcover are re-projected from 30m), centered on observation location.
The data were extracted from rasters publicly available at \href{https://stac.ecodatacube.eu/}{Ecodatacube}, \href{https://lpdaac.usgs.gov/products/astgtmv003/}{ASTER}, and \href{https://worldcover2021.esa.int/}{ESA WorldCover}. The data are available in the form of torch tensors in a shape of [$6\times64\times64$].

\begin{figure}[h]
    \centering
    \includegraphics[width=0.235\linewidth]{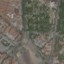} 
    \includegraphics[width=0.235\linewidth]{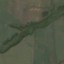}
    \includegraphics[width=0.235\linewidth]{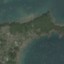}
    \includegraphics[width=0.235\linewidth]{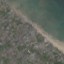} 
    \caption{\textbf{Satellite RGB images} with 64$\times$64 resolution extracted from Sentinel-2A rasters available at \href{https://stac.ecodatacube.eu/}{Ecodatacube}. }
    \label{fig:satellite_images}
\end{figure}

\noindent\textbf{Meteorological data} and other climatic variables are vital assets for species identification and distribution modeling \cite{beaumont2005predicting, hijmans2006ability}. In light of that, we provide 20 years of historical time-series monthly values of mean, min., and max. temperature and total precipitation for all observations (see Figure \ref{fig:temperature_graph} in Appendix for example data). For each observation site, 20 years of data was extracted; for instance, an observation from 2000 includes data from 1980 to 2000. However, as the available climatic rasters only extend up to the year 2020, observations from 2020 to 2024 have missing values for those years not covered by existing data. 
In addition, we provide 19 bioclimatic variables (e.g., temp., seasonality, etc.) averaged over the period from 1981 to 2010. All data were extracted from \href{https://chelsa-climate.org/bioclim/}{CHELSA} \cite{karger2017climatologies1,karger2017climatologies2}. \\

\begin{figure}[h]
\footnotesize
\begin{tabular}{@{\hspace{1mm}}p{2.1cm}@{\hspace{1mm}}p{6cm}@{}}

\includegraphics[width=2cm]{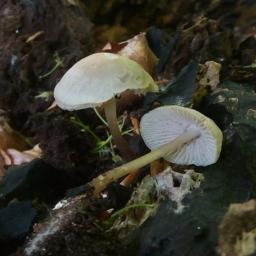} & \vspace{-2cm} 
        \textit{Its stem is thick and light brown, with a hint of green at the base.
        The smaller mushroom on the right has a similar light brown cap, but its rim is more pronounced and has a white, almost translucent appearance. This gives it a delicate, lacy look. The stem of this mushroom is thinner and lighter in color compared to ......}
\end{tabular}
\vspace{-1mm}
\caption{\textbf{Image caption sample.} For each photograph, we use a Malmo-7B \cite{deitke2024molmo} VLM to produce a realistic image caption with an exhaustive text description.}
\label{fig:image-captioning}
\vspace{-2mm}
\end{figure}

\noindent\textbf{Image captions}. Recent advances in VLMs \cite{li2023blip,Achiam2023GPT4TR,alayrac2022flamingo} have demonstrated strong performance across tasks such as image reasoning \cite{Achiam2023GPT4TR} and captioning \cite{li2023blip} and shown that VLMs can effectively understand and reason about fine-grained details within images \cite{liu2024democratizing}. 
Building on these insights, we provide text descriptions for most photographs using the state-of-the-art open-source Malmo-7B VLM model~\cite{deitke2024molmo}. We generate baseline captions (see Figure \ref{fig:image-captioning} and Figure \ref{fig:image-captioning2}, App.) with a prompt specifically designed to emphasize visual characteristics relevant to fungi identification, while avoiding unnecessary or potentially misleading details. The following prompt was used to guide the caption generation:
\begin{center}
\vspace{-1mm}
\begin{minipage}{0.95\linewidth}
\small
\textit{``Describe the visual features of the fungi, such as their colour, shape, texture, and relative size. Focus on the fungi and their parts. Provide a detailed description of the visual features, but avoid speculations.''}
\end{minipage}
\end{center}

\noindent\textbf{Location-related metadata} is provided for approximately 99.9\% of the observations and describes the location, time, taxonomy, and toxicity of the specimen, surrounding environment, and capturing device.
See Table \ref{tab:data_attributes} for a detailed description of all available location-related metadata.
While part of the metadata is usually provided by citizen scientists\footnote{A member of the public who actively participates in data collection, contributing valuable information to support professional scientists.}, some attributes (e.g., elevation, land cover, and biogeographical) are crawled externally; all with potential to improve the classification accuracy and enable research on combining visual data with metadata. \vspace{-3mm}\\

\begin{table}[h]
\centering
\renewcommand*\arraystretch{1.3}
    \caption{\textbf{List of available location-related metadata}. For virtually all observations (>99.9\%), we provide data describing the surroundings or the specimen. Using such data for species identification allows to improve accuracy; see \cite{diao2022metaformer,picek2022danish}.
\label{tab:data_attributes}
}
\begin{tabular}{@{}p{1.7cm}p{6.1cm}@{}}
\toprule
\small{\textbf{Metadata}} & \small{\textbf{Description}} \\
\midrule
\textbf{\footnotesize{Observation date}} & \footnotesize{Date when the specimen was observed in yyyy-mm-dd format. Besides, three additional columns with pre-extracted \textit{year}, \textit{month}, and \textit{day}} values are provided. \\
\textbf{\footnotesize{EXIF}} & \footnotesize{Camera attributes extracted from the image, e.g., metering mode, color space, device type, exposure, etc.} \\
\textbf{\footnotesize{Habitat}} & \footnotesize{The environment where the specimen was observed. Selected from 32 values such as \textit{Mixed woodland}, \textit{Deciduous woodland}, etc.} \\
\textbf{\footnotesize{Substrate}} & \footnotesize{The natural substance on which the specimen lives. Selected from 32 values such as \textit{Bark}, \textit{Soil}, \textit{Stone}, etc.} \\
\textbf{\footnotesize{Taxonomic   \,\,labels}} & \footnotesize{For each observation, we provide full taxonomic labels that include all ranks from species level up to kingdom. All are available in separate columns.} \\
\textbf{\footnotesize{Toxicity status}} & \footnotesize{Whether the species is poisonous or not as a binary value. Since non-edible species can cause serious health issues as well, we label them as poisonous.} \\
\textbf{\footnotesize{Location}} & \footnotesize{Latitude + longitude and coarser administrative divisions into regions, districts, and countries.} \\
\textbf{\footnotesize{Biogeographical zone}} & \footnotesize{One of the major biogeographical zones, e.g., \textit{Atlantic}, \textit{Continental}, \textit{Alpine}, \textit{Mediterranean}, and \textit{Boreal}.} \\
\textbf{\footnotesize{Elevation}} & \footnotesize{Standardized height above the sea level.} \\ \textbf{\footnotesize{Landcover}} & \footnotesize{Land cover classification code with values like \textit{savanna}, \textit{barren}, etc. Taken from MODIS Terra+Aqua \cite{friedl2010modis}.} \\
\bottomrule
\end{tabular}
\vspace{-0.3cm}
\end{table}

\begin{figure*}[!h]
    \centering
    \includegraphics[width=\textwidth]{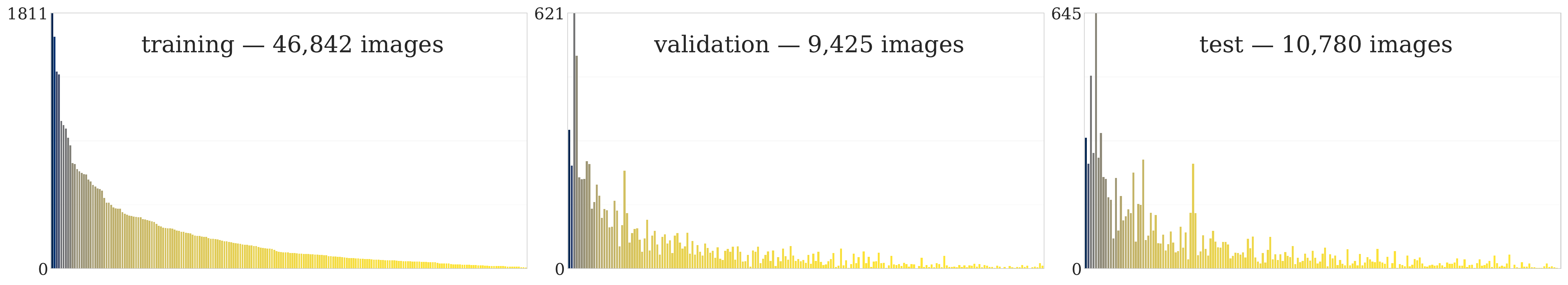}
    \caption{\textbf{Class distribution shift on the \ac{fungim} dataset}.The long-term data acquisition captures a phenomenon related to natural changes in species presence, i.e., class prior shift. Sorted in descending order based on their occurrence in the training set. The training set includes data from 2021 and before (215 species), the validation set from 2022 (196 species), and the test set from 2023 (193 species).}
    \label{fig:mini_data_distribution}
    \vspace{-0.25cm}
\end{figure*}

\section{FungiTastic Benchmarks}

\label{sec:challenges}

The diversity and unique features of the FungiTastic dataset allow for the evaluation of various fundamental computer vision and machine learning problems. 
We present several benchmarks, each with its own evaluation protocol. This section provides a detailed description of each challenge and the corresponding evaluation metrics. Metrics are further defined in Appendix \ref{app:eval_metrics}. \\

\noindent\textbf{Closed-set classification}:
The FungiTastic dataset is a challenging dataset with many visually similar species, a heavy long-tailed distribution, and considerable distribution shifts over time. Since the fine-grained closed-set classification methodology is well-defined, we follow the widely accepted standard, and we, apart from accuracy, use the macro-averaged F1-score ($\text{F}_{1}^{m}$). \\

\noindent\textbf{Open-set classification}:
In the Atlas of Danish Fungi (our data source), new species are continuously added to the database, including previously unreported species. This long-term ongoing data acquisition enables a yearly data split with a natural class distribution shift (see Figure \ref{fig:mini_data_distribution}), and many species in the test data are absent in the training set. We follow a widely accepted methodology, and we propose to use an AUC as the main metric. Besides, we calculate True Negative Rate at 95\% True Positive Rate (TNR$^{95}$) metric. \\

\noindent\textbf{Few-shot classification}: All the categories with less than five samples, usually uncommon and rare species, form the few-shot subset. Being capable of recognizing those is of high interest to the experts. Since the few-shot dataset has no severe class imbalance like the other FungiTastic subsets, the main metric is Top1 accuracy. The macro-averaged F1-score ($\text{F}_{1}^{m}$) and Top3 total accuracy are also reported. This challenge does not have any ``unknown'' category. \\

\noindent\textbf{Chronological classification}:
Each observation in the FungiTastic dataset has a timestamp, allowing the study of species distribution changes over time. Fungi distribution is seasonal and influenced by weather, such as recent precipitation. New locations may be added over time, providing a real-world benchmark for domain adaptation methods, including online, continual, and test-time adaptation. The test dataset consists of fungi images ordered chronologically, meaning a model processing an observation at time $t$ can access all observations with timestamps $t' < t$. \\

\noindent\textbf{Classification beyond 0--1 loss function}:
Evaluation of classification networks is typically based on the 0--1 loss function, such as the mean accuracy, which also applies to the metrics defined for the previous challenges. This often falls short of the desired metric in practice since not all errors are equal. 
In this challenge, we define two practical scenarios: In the first scenario, confusing a poisonous species for an edible one (false positive edible mushroom) incurs a much higher cost than that of a false positive poisonous mushroom prediction. In the second scenario, the cost of not recognizing that an image belongs to a new species should be higher. \\

\noindent\textbf{Segmentation}: Acquiring human-annotated segmentation masks can be resource-intensive, yet segmentation is vital for advanced recognition and fine-grained classification methods \cite{bhatt2024mitigating, rigotti2021attention}. Accurate segmentation of fungal images supports these methods and enables automated analysis of species-specific morphological and environmental relationships and revelation of ecological and morphological patterns across locations. With its annotations, FungiTastic-M is built to accommodate semantic segmentation using the standard mean Intersection over Union (mIoU) metric and instance segmentation with the mean Average Precision (mAP) metric.

\section{Baseline Experiments}
\label{sec:experiments}
In this section, we describe various weak and strong baselines based on state-of-the-art architectures and methods for four \acs{fungi} benchmarks. 
We report results for the closed-set, few-shot learning, and zero-shot segmentation, but other baselines will be provided later in the supplementary materials, documentation, or on the dataset website.

\subsection{Closed-set Image Classification}
\label{subsec:exp:closed_set}

We train a variety of state-of-the-art CNN architectures to establish some baselines for closed-set classification on the FungiTastic and \ac{fungim}. All selected architectures were optimized with Stochastic Gradient Descent with momentum set to 0.9, SeeSaw loss \cite{wang2021seesaw}, a mini-batch size of 64, and Random Augment \cite{cubuk2020randaugment} with a magnitude of 0.2. The initial LR was set to 0.01 (except for ResNet and ResNeXt, with LR=0.1), and it was scheduled based on validation loss.

\textbf{Results:} Similarly to other fine-grained benchmarks, while the number of params, complexity of the model, and training time are more or less the same, the transformer-based architectures achieved considerably better performance on both FungiTastic and \ac{fungim} and two different input sizes (see Table \ref{tab:results_closed_set} and Table \ref{tab:results_closed_set2} in Appendix). The best-performing model, BEiT-Base/p16 \cite{bao2021beit}, achieved $\text{F}_{1}^{m}$ just around 40\%, which shows the severe difficulty.

\begin{table}[!h]
\footnotesize
\begin{center}
\setlength{\tabcolsep}{0.6em} %
\renewcommand{\arraystretch}{1.1}
\caption{
\textbf{Closed-set fine-grained classification on \ac{fungi} and \ac{fungim}}.
A set of selected state-of-the-art Convolutional- (top section) and Transformer-based (bottom section) architectures evaluated on test sets. All reported metrics show the challenging nature of the dataset.
}
\label{tab:results_closed_set}
\begin{tabular}{@{}l| c  c  c | c  c  c@{}}
\toprule
  & \multicolumn{3}{c|}{\scriptsize \ac{fungim} -- 224$^2$} & \multicolumn{3}{c}{\scriptsize \ac{fungi} -- 224$^2$}  \\  
    \multicolumn{1}{@{}l|}{\textit{Architecture}} & \textbf{Top1} & \textbf{Top3} & \,\,\textbf{$\text{F}_{1}^{m}$}\,\, & \textbf{Top1} & \textbf{Top3} & \,\,\textbf{$\text{F}_{1}^{m}$}\,\, \\  
	\midrule
        ResNet-50 \cite{he2016deep}               & 61.7 & 79.3 & 35.2 & 62.4 & 77.3 & 32.8 \\
	ResNeXt-50 \cite{xie2017aggregated}             & 62.3 & 79.6 & 36.0 & 63.6 & 78.3 & 33.8 \\
	EfficientNet-B3 \cite{tan2019efficientnet}        & 61.9 & 79.2 & 36.0 & 64.8 & 79.4 & 34.7 \\
	EfficientNet-v2-B3 \cite{tan2021efficientnetv2}  & 65.5 & 82.1 & 38.1 & 66.0 & 80.0 & 36.0 \\
        ConvNeXt-Base \cite{liu2022convnet}          & 66.9 & 84.0 & 41.0 & 67.1 & 81.3 & 36.4 \\
        \midrule
        ViT-Base/p16 \cite{dosovitskiy2020image}           & 68.0 &  \underline{84.9} & 39.9 & \underline{69.7} &  \underline{82.8} & \underline{38.6}  \\
        Swin-Base/p4w12 \cite{liu2021swin}        & \textbf{69.2} & \textbf{85.0} & 42.2 & 69.3 & 82.5 & 38.2  \\
 	BEiT-Base/p16 \cite{bao2021beit}          & \underline{69.1} & 84.6 & \textbf{42.3} & \textbf{70.2} & \textbf{83.2} & \textbf{39.8} \\
   \bottomrule
\end{tabular}
\end{center}
\vspace{-0.5cm}
\end{table}

\subsection{Few-shot Image Classification}

Three baseline methods are implemented. The first baseline is standard classifier training with the \ac{ce} loss. The other two baselines are nearest-neighbor classification and centroid prototype classification based on deep embeddings extracted from large-scale pre-trained vision models, namely CLIP \cite{radford2021learning}, BioCLIP \cite{stevens2024bioclip} and DINOv2\,\cite{oquab2023dinov2}.

Standard deep classifiers are trained with the \ac{ce} loss to output the class probabilities for each input sample.
Nearest neighbors classification ($\mathbf{k}$-NN) constructs a database of training image embeddings. At test time, $k$ nearest neighbors are retrieved, and the classification decision is made based on the majority class of the nearest neighbors.
Nearest-centroid-prototype classification constructs a prototype embedding for each class by aggregating the training data embeddings of the given class. The classification depends on the image embedding similarity to the class prototypes. These methods are inspired by prototype networks proposed in\,\cite{snell2017prototypical}. 

\textbf{Results:} While DINOv2 \cite{oquab2023dinov2} embeddings greatly outperform CLIP \cite{radford2021learning} embeddings, BioCLIP \cite{stevens2024bioclip} outperforms them both, highlighting the dominance of domain-specific models. Further, the centroid-prototype classification always outperforms the nearest-neighbor methods. Finally, the best standard classification models trained on the in-domain few-shot dataset underperform both DINOv2 and CLIP embeddings, which shows the power of methods tailored to the few-shot setup. For results summary, refer to Table \ref{tab:results_few_shot}.

\begin{table}[h]
    \caption{\textbf{Few shot classification on \acl{fungif}}. Pretrained deep descriptors with the nearest centroid and 1-NN nearest neighbor classification (Left) and fully supervised  (max 4 examples per class) classifier with cross-entropy-loss (Right).
    All pre-trained models are based on the ViT-B architecture, CLIP \cite{radford2021learning}, and BioCLIP \cite{stevens2024bioclip} with patch size 32 and DINOv2 \cite{oquab2023dinov2} with patch size~16.}
    \vspace{-0.1cm}
    \centering
    \footnotesize
    \renewcommand{\arraystretch}{1.1}
    \hspace{-1.1cm}
    \begin{minipage}{0.6\linewidth}
        \begin{tabular}{@{}l@{\hspace{5pt}}l@{\hspace{4pt}}|@{\hspace{4pt}}c@{\hspace{4pt}}c@{}}
        \toprule
        \textbf{Model} & \textbf{Method}  & \textbf{Top1}  & \textbf{Top3}\\
        \midrule
        \multirow[c]{2}{*}{CLIP}  & 1-NN & 6.1 & -- \\
                                  & centroid & 7.2 & 13.0  \\
        \midrule
        \multirow[c]{2}{*}{DINOv2} & 1-NN & 17.4 &  -- \\
                                   & centroid & 17.9 &  27.8 \\
                                   
        \midrule
        \multirow[c]{2}{*}{BioCLIP} & 1-NN & 18.8  & -- \\
                                    & centroid & \textbf{21.8} &  \textbf{32.6}  \\
        \bottomrule
        \end{tabular}
    \end{minipage}
    \hspace{-1.2cm}
    \begin{minipage}{0.385\linewidth}
        \begin{tabular}{@{}l@{\hspace{5pt}}c@{\hspace{4pt}}|@{\hspace{4pt}}c@{\hspace{4pt}}c@{}}
        \toprule
        \textbf{Architecture} & \textbf{Input}  & \textbf{Top1} & \textbf{Top3}\\
        \midrule
        \multirow[c]{2}{*}{BEiT-B/p16} & {\scriptsize 224$\times$224} & 11.0 & 17.4  \\
                                       & {\scriptsize 384$\times$384} & 11.4 & 18.4  \\
        \midrule
        \multirow[c]{2}{*}{ConvNeXt-B} & {\scriptsize 224$\times$224} & 14.0 & 23.1  \\
                                       & {\scriptsize 384$\times$384} & 15.4 & 23.6  \\
            
        \midrule
        \multirow[c]{2}{*}{ViT-Base/p16}  & {\scriptsize 224$\times$224} & 13.9 & 21.5 \\
                                       & {\scriptsize 384$\times$384} & 19.5 & 29.0  \\
        \bottomrule
        \end{tabular}
    \end{minipage}
    
    \vspace{-0.25cm}
    \label{tab:results_few_shot}
\end{table}

\subsection{Experiments with Additional Metadata} 
We provide baseline experiments using tabular metadata (habitat, month, substrate) based on previous work \cite{picek2022danish}. Table \ref{tab:exps:tab_data} shows that all the attributes improve all the metrics. Individually, the addition of the habitat attribute results in the biggest gains in accuracy (2.3\%), followed by substrate (1.2\%) and month (0.9\%). 
Overall, habitat was the most efficient way to improve performance. With the combination of Habitat, Substrate, and Month, we improved the EfficientNet-B3 model’s performance on FungiTastic-M by 3.62\%, 3.42\% and 7.46\% in Top1, Top3, and F1, respectively, indicating the gains are mostly orthogonal. 
Using the MetaSubstrate instead of  Substrate resulted in performance lower by 0.2\%, 0.5\%, and 0.3\% in Top1, Top3, and F1, respectively.

\begin{table}[h]
\vspace{-0.15cm}
\footnotesize
\setlength{\tabcolsep}{0.3em} %
\renewcommand{\arraystretch}{1.17}
\centering
\caption{\textbf{Ablation on a combination of observation-related data.} Utilizing a simple yet effective approach based on previous work \cite{picek2022danish}, we measure performance improvement using Habitat, Substrate, and Month and their combination. We also test how replacing Substrate variables with MetaSubstrate affects performance. Evaluated with EfficientNet-B3 on FungiTastic-M test set.
}
\vspace{-0.1cm}
\begin{tabular}{@{}l@{\hspace{-0.4mm}}cccc|ccccc|cc@{}}
\toprule
\textit{Habitat}          & \checkmark & --  & --   & --  & \checkmark & \checkmark      & \checkmark & --  & --     & \checkmark & \checkmark          \\
\textit{Month}          & --  & \checkmark & --  & --   & \checkmark & --  & --   & \checkmark & \checkmark        & \checkmark & \checkmark          \\ 
\textit{Substrate}        & --  & --   & \checkmark & --  & --   & \checkmark & --  & \checkmark & --  & \checkmark      & --      \\ 
\textit{MetaSub.}   & --  & --   & --  & \checkmark & --  & --   & \checkmark & --  & \checkmark      & --  & \checkmark         \\ \midrule
\textbf{Top1}    & \textbf{+2.3}    & +0.9   & \underline{+1.2}   & +0.9   & +3.1   & +3.0   & +2.7   & +1.9   & +1.6  & +3.6   & +3.3    \\ 
\textbf{$\text{F}_{1}^{m}$}     & \textbf{+4.0}    & +1.1   & \underline{+2.3}   & +1.5   & +6.0   & +5.9   & +5.1   & +4.0   & +3.2  & +7.5   & +6.8   \\ 
\textbf{Top3}   & \textbf{+2.3}    & +0.5   & \underline{+0.8}   & +0.6   & +2.7   & +2.9   & +2.6   & +1.5   & +1.1  & +3.4   & +3.1    \\ 
\bottomrule
\label{tab:exps:tab_data}
\end{tabular}
\vspace{-0.5cm}
\end{table}

\subsection{Segmentation}
A zero-shot baseline for foreground-background binary segmentation of fungi is evaluated on the \ac{fungim} dataset. The method consists of two steps: 1. The GroundingDINO \cite{liu2023grounding} (the `tiny' version of the model) zero-shot object detection model is prompted with the text `mushroom' and outputs a set of instance-level bounding boxes. 2. The bounding boxes from the first step are used as prompts for the SAM \cite{Kirillov_2023_ICCV} segmentation model. All the experiments are conducted on images with the longest edge resized to 300 pixels while preserving the aspect ratio.

\begin{figure}[!h]
\begin{center}
     \includegraphics[height=1.45cm]{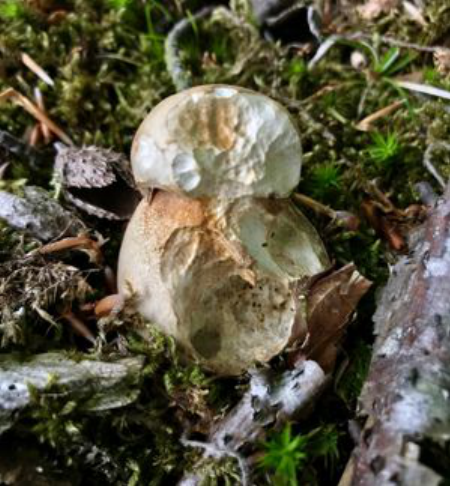}
     \includegraphics[height=1.45cm]{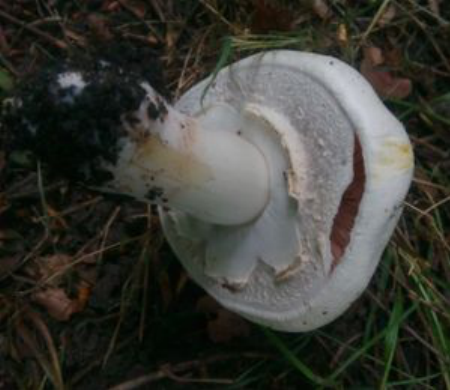}
     \includegraphics[height=1.45cm]{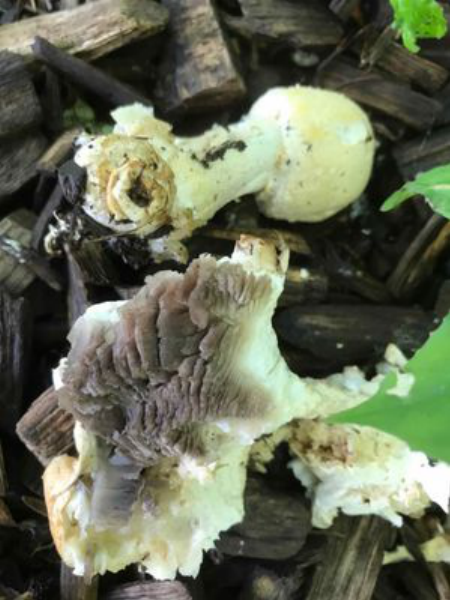}
     \includegraphics[height=1.45cm]{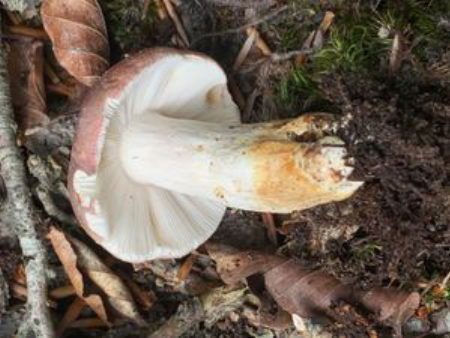}
     \includegraphics[height=1.45cm, width=1.93cm]{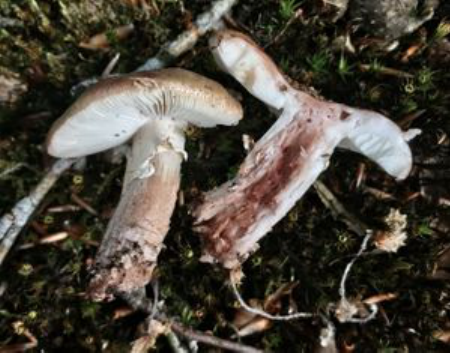} \\
     \vspace{1px}
     \includegraphics[height=1.45cm]{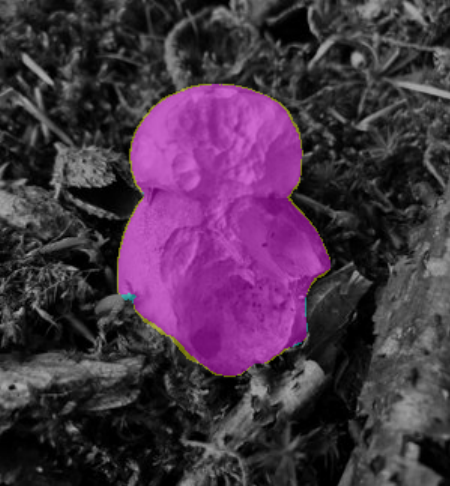}
     \includegraphics[height=1.45cm]{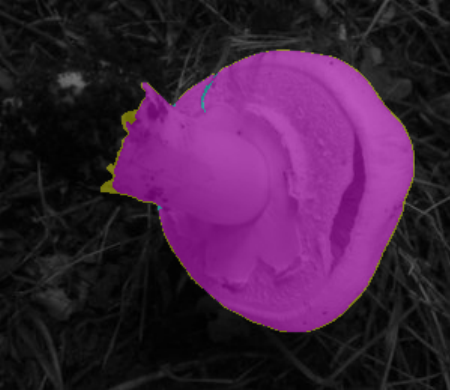}
     \includegraphics[height=1.45cm]{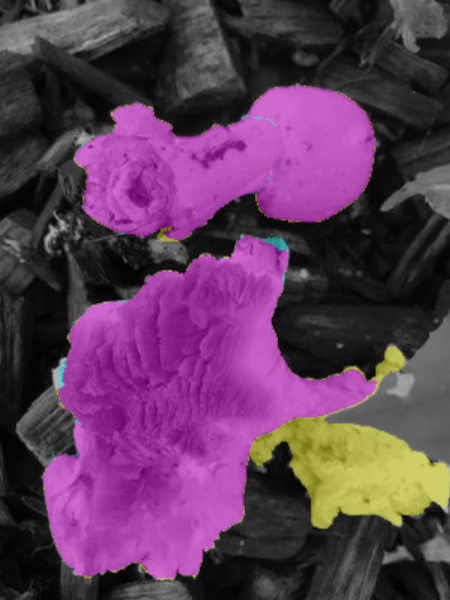}
     \includegraphics[height=1.45cm]{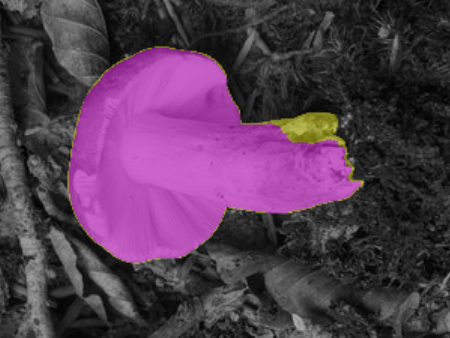}
     \includegraphics[height=1.45cm, width=1.93cm]{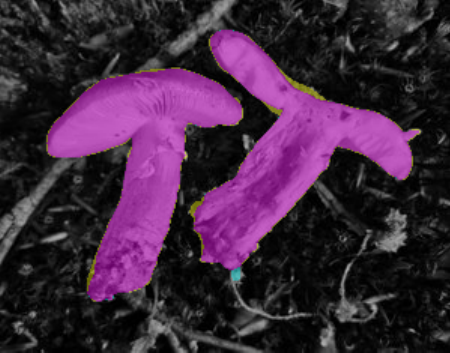} \\
             \vspace{10px}
     \includegraphics[height=1.85cm]{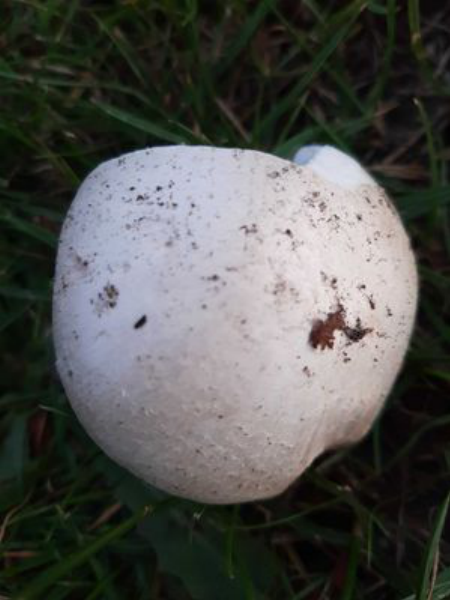}
     \includegraphics[height=1.85cm]{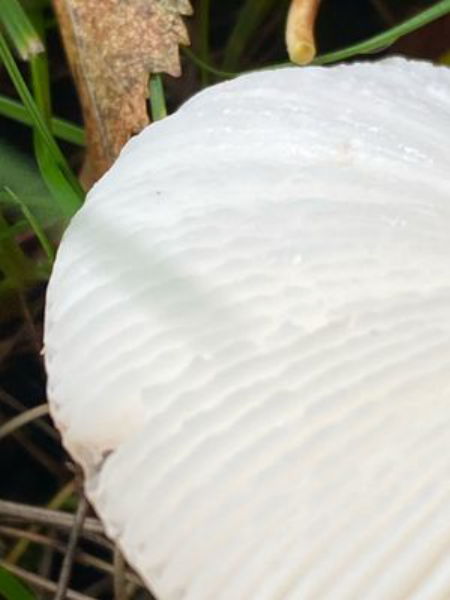}
     \includegraphics[height=1.85cm]{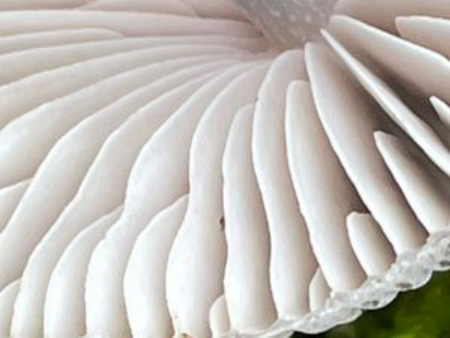}
     \includegraphics[height=1.85cm]{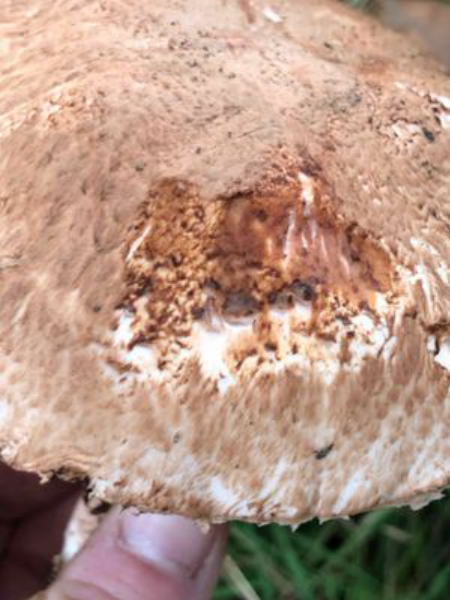} 
     \includegraphics[height=1.85cm, width=1.3cm]{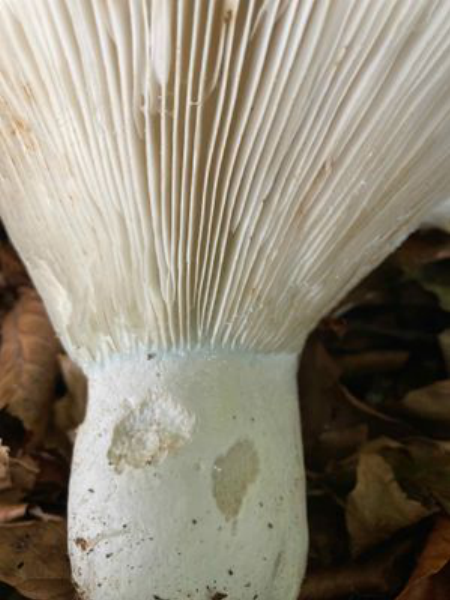} \\
        \vspace{1px}
     \includegraphics[height=1.85cm]{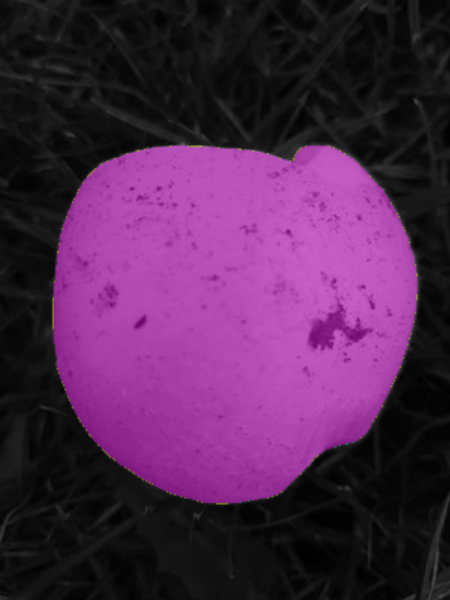}
     \includegraphics[height=1.85cm]{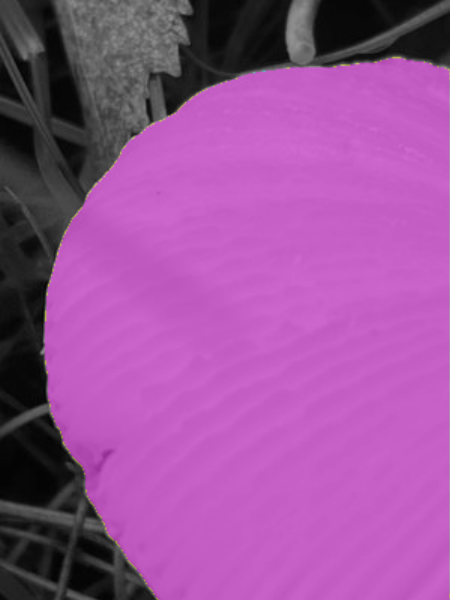}
     \includegraphics[height=1.85cm]{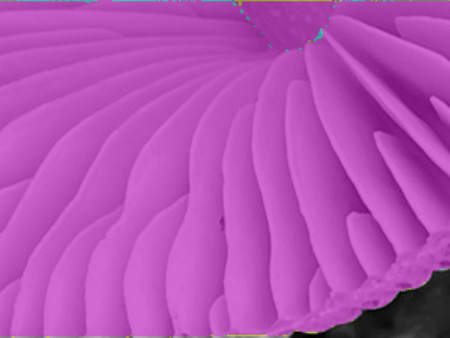}
     \includegraphics[height=1.85cm]{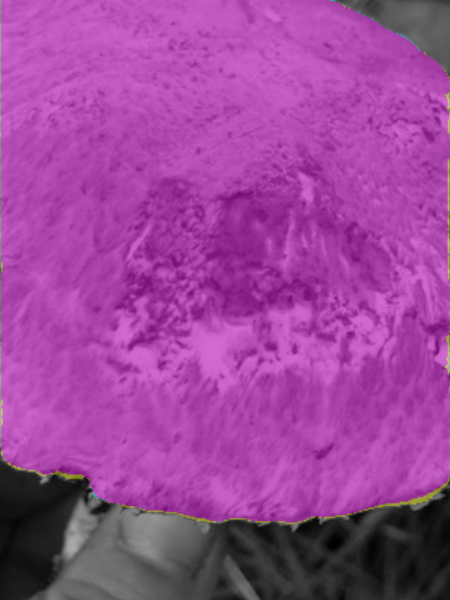}
     \includegraphics[height=1.85cm, width=1.3cm]{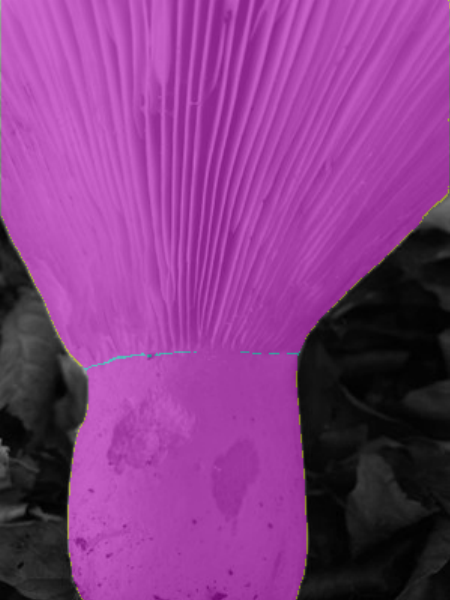} \\
        \vspace{10px}
     \includegraphics[height=1.45cm]{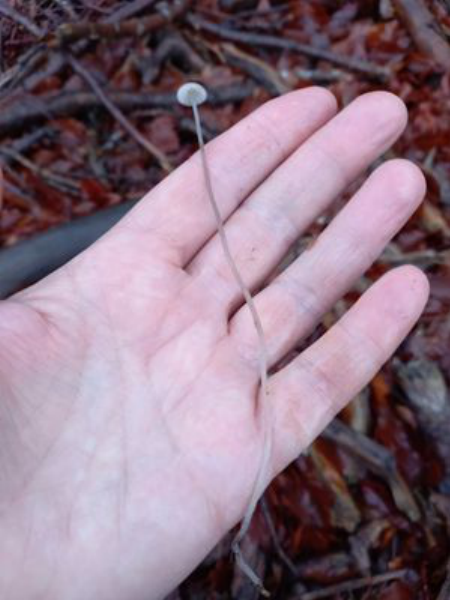}
     \includegraphics[height=1.45cm]{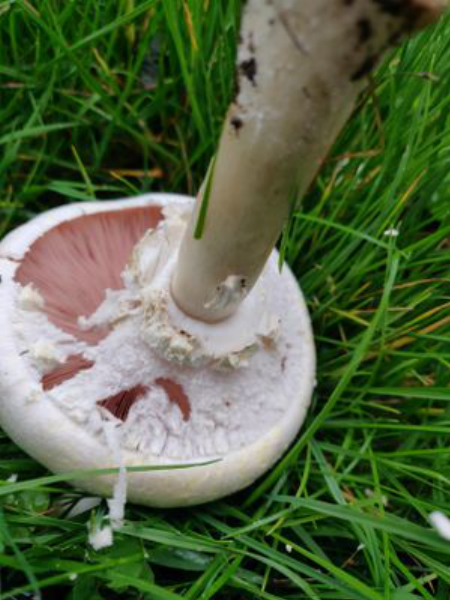}
     \includegraphics[height=1.45cm]{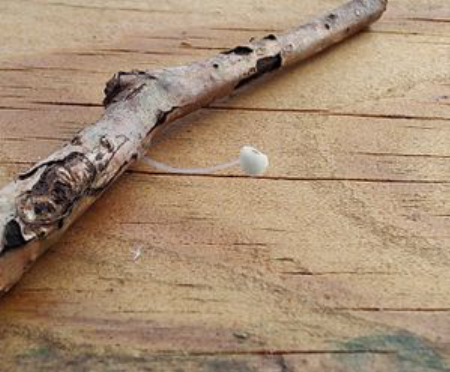}
     \includegraphics[height=1.45cm, width=2.05cm]{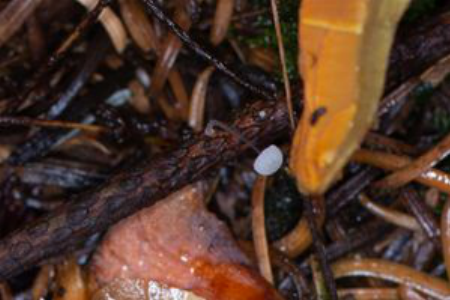}
     \includegraphics[height=1.45cm]{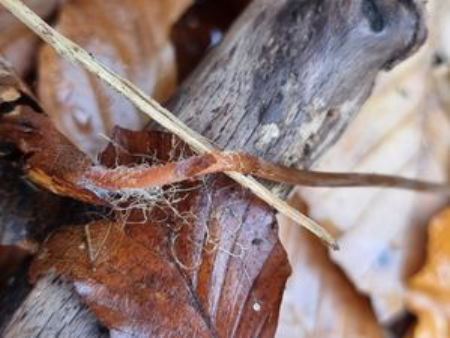} \\
     \vspace{1px}
     \includegraphics[height=1.45cm]{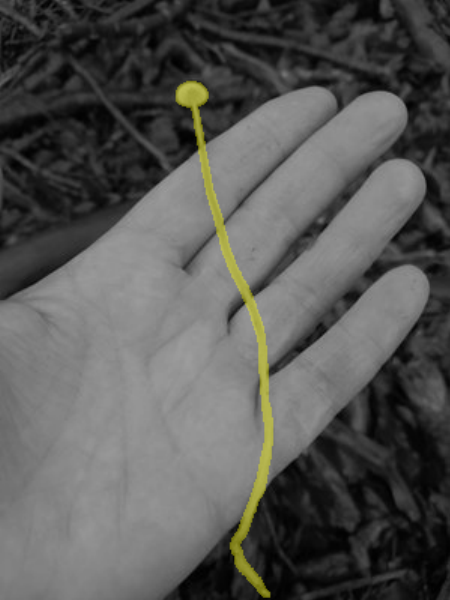}
     \includegraphics[height=1.45cm]{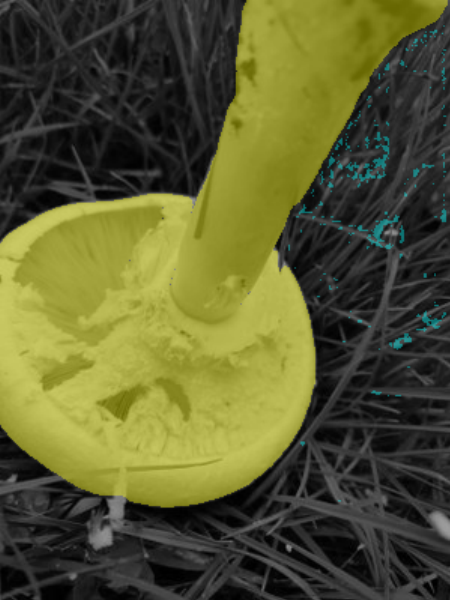}
     \includegraphics[height=1.45cm]{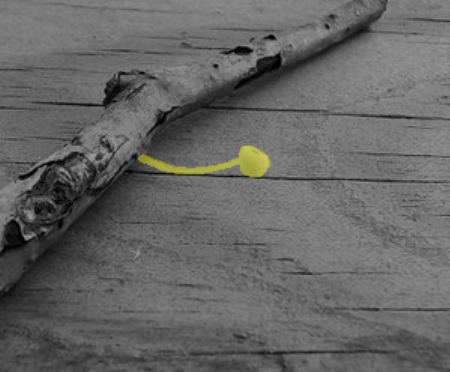}
     \includegraphics[height=1.45cm, width=2.05cm]{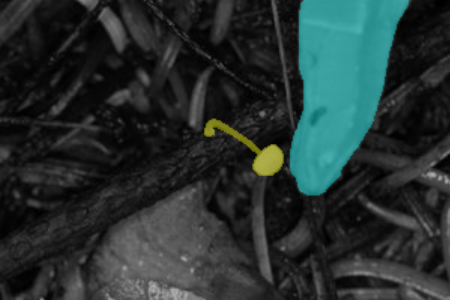}
     \includegraphics[height=1.45cm]{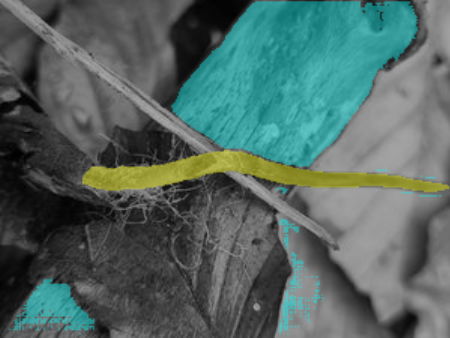} \\
\vspace{-0.5cm}
\end{center}
\caption{\textbf{Zero-shot Fungi segmentations on \ac{fungim} benchmark}.
Random samples (top section), best IoU samples (mid), and worst IoU samples (bottom).
 Highlighted pixels correspond to:
  {\setlength{\fboxsep}{1pt}\colorbox{magenta!30}{true positives}},
  {\setlength{\fboxsep}{1pt}\colorbox{cyan!30}{false positives}}, and
  {\setlength{\fboxsep}{1pt}\colorbox{yellow!30}{false negatives}}.}
\label{fig:seg_results}
\end{figure}

\textbf{Results:} The baseline method achieved an average per-image IoU of 89.36\%. While the model exhibits strong zero-shot performance, it sometimes fails to detect mushrooms. These instances often involve small mushrooms, where a higher input resolution could enhance detection, and atypical mushrooms, such as very thin ones. Another common issue is SAM’s tendency to miss mushroom stems. The results for the simplified foreground-background segmentation task underline the need for further development of domain-specific models. Qualitative results, including random images and examples where the segmentation performs best and worst, are reported in Figure \ref{fig:seg_results}.

\subsection{Open-set Image Classification}
\label{sec:openset}
While constructing baselines for the FungiTastic open-set benchmark, we approached open-set classification as a binary decision-making problem, where the model determines whether a new image belongs to a known class or a novel class. This method serves as an initial step in the classification pipeline, deciding if a closed-set classifier is suitable for recognizing a given sample. We evaluate several approaches for open-set classification:
\begin{itemize}
    \item Maximum Softmax Probability \cite{hendrycks2017a} (MSP): Uses the highest probability from softmax output from a closed-set classifier as the open-set score.
    \item Maximum Logit Score \cite{vaze2022openset} (MLS): Uses the highest logit output from a closed-set classifier as the open-set score.
    \item Nearest Mean Score (NM): Computes the mean embedding for each class, then calculates the Euclidean distance between an image embedding and the nearest class mean.
\end{itemize}

We use features and logits from the BEiT-Base/p16 closed-set classifier baseline, trained on the full dataset (i.e., FungiTastic) at a 384$\times$384 resolution. To explore the potential of general pre-trained representation, we compare the fully-supervised model with generic features from a pre-trained DINOv2 model \cite{oquab2023dinov2}. Using DINOv2 features, we train a simple linear layer to obtain MSP and MLS scores. Note that BEiT represents the best model from the closed-set classification baseline.

\textbf{Results}: The MLS method achieved the best open-set classification performance on both backbones. With BEiT-Base/p16, MLS achieves a TNR$^{95}$ of 27.7\% and an AUC of 83.9\%, which are the highest AUC across all methods. DINOv2, in contrast, achieves the best TNR$^{95}$ with a score of 36.9\% using the MLS method, though its AUC is slightly lower at 74.5\%. The MSP method also performs well with DINOv2, reaching a TNR$^{95}$ of 32.5\% and an AUC of 82.4\%. However, the NM method, which relies on feature embeddings rather than classifier outputs, significantly underperforms in both metrics. See Table \ref{table:openset} for more details.

\begin{table}[h]
\footnotesize
\setlength{\tabcolsep}{0.6em} %
\centering
\caption{\textbf{Open-set classification baselines.} Overall, the results are inconclusive and highly metric-dependent. The MLS (\textit{Max. Logit}) method with the BEiT-Base/p16 backbone yields the highest AUC (83.9\%), while the DINOv2 backbone with MLS achieves the highest TNR$^{95}$ (36.9\%). The NM (\textit{Nearest Mean}) method consistently underperforms in both metrics across both backbones. For the AUC metric, MLS with a BEiT backbone (fine-tuned on the FungiTastic closed-set dataset) outperforms other approaches. However, both MSP (\textit{Max. Softmax}) and MLS using DINOv2 linear layer are better when TNR$^{95}$ performance is considered.
}
\begin{tabular}{@{}lcc|cc|cc@{}}
    \toprule
     & \multicolumn{2}{c|}{\textit{Nearest Mean}} & \multicolumn{2}{c|}{\textit{Max. Logit}} & \multicolumn{2}{c@{}}{\textit{Max. Softmax}} \\
    \textit{Backbone} & \textbf{TNR$^{95}$} & \textbf{ AUC} & \textbf{TNR$^{95}$} & \textbf{AUC} & \textbf{TNR$^{95}$} & \textbf{AUC} \\
    \midrule
    BEiT-Base/p16   & 23.2 & 73.9 & 27.7 & 83.9 & 25.3 & 79.8 \\
    DINOv2          & 12.1 & 69.2 & 36.9 & 74.5 & 32.5 & 82.4 \\
    \bottomrule
\end{tabular}
\label{table:openset}
\end{table}

\subsection{Vision-Language Fusion}
To evaluate the relevance of the available textual data (i.e., photograph captions) for species classification, we provide baselines that use a sequence classification variant of the lightweight DistilBERT \cite{Sanh2019DistilBERTAD} model trained as a classifier on textual descriptions only. The model was trained for 10 epochs using the standard cross-entropy loss, with logits obtained from a classification head applied to the pooled features of the class token in DistilBERT. For evaluation, we use text descriptions generated for the images in the test set.

\begin{figure}[hb]
\vspace{-0.35cm}
    \centering
    \includegraphics[width=1.0\linewidth]{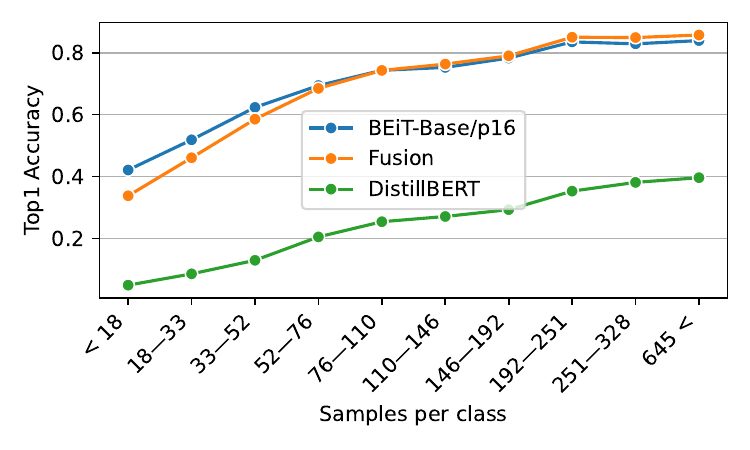}
    \vspace{-0.25cm}
    \caption{\textbf{Vision-Language fusion -- accuracy dependence on class frequency}.
    Like the vision model (BEiT-Base/p16), the language model (DistillBERT) struggles with infrequent classes. Fusion improves accuracy mainly for species with over 100 samples. The test set is binned by class frequency into deciles (x-axis).}
    \label{fig:samples_cls}
\end{figure}

\textbf{Results:} The DistilBERT classifier achieves a Top1 accuracy of 31.2\% on FungiTastic-M and 24.1\% on the full benchmark; significantly lower than fully supervised BEiT classifiers. However, this is still a strong result, given that it relies solely on textual descriptions. A simple ensemble that averages logits from the image and text classifiers shows potential for improved accuracy, indicating that the two methods are complementary. VLM-based descriptions capture useful details often missed by the image model. Further analysis shows the ensemble improves performance mainly on common categories, while the image classifier performs better on rare ones. This trade-off likely accounts for the drop in $\text{F}_{1}^{m}$ and overall accuracy on the full benchmark. For more details, see Table \ref{tab:results_vl} and Figure \ref{fig:samples_cls}.

\begin{table}[hb]

\footnotesize
\begin{center}
\setlength{\tabcolsep}{0.6em}
\caption{
\textbf{Vision-Language fusion performance}. DistillBERT uses text descriptions of images for species classification. Fusion method predictions are the mean of DistillBERT and BEiT logits.}
\label{tab:results_vl}
\begin{tabular}{@{}l| c  c  c | c  c  c@{}}
\toprule
  & \multicolumn{3}{c|}{\scriptsize \ac{fungim} -- 224$^2$} & \multicolumn{3}{c@{}}{\scriptsize \ac{fungi} -- 224$^2$} \\  
    \multicolumn{1}{@{}l|}{\textit{Architectures}} & \textbf{Top1} & \textbf{Top3} & \,\,\textbf{$\text{F}_{1}^{m}$}\,\, & \textbf{Top1} & \textbf{Top3} & \,\,\textbf{$\text{F}_{1}^{m}$}\,\, \\  
	\midrule
    DistillBERT     & 31.2 & 50.2 & 11.5 & 24.1 & 39.1 & 8.8 \\
 	BEiT-Base/p16   & \underline{67.3} & \underline{83.3} & \textbf{40.5} & \textbf{70.2} & \textbf{83.2} & \textbf{41.1} \\
    \midrule
    Fusion          & \textbf{67.7} & \textbf{83.8} & \underline{39.8} & \underline{69.0} & \underline{82.6} & \underline{40.0} \\
   \bottomrule
\end{tabular}
\end{center}
\vspace{-0.35cm}
\end{table}
    
\section{Conclusion}
\label{sec:conclusion}
In this work, we introduced the \acl{fungi}, a comprehensive and multi-modal dataset and benchmark. The dataset includes a variety of data types, such as photographs, satellite images, climatic data, segmentation masks, and observation metadata. 
\acl{fungi}  has many interesting features, which make it attractive to the broad ML community. With its data sampling spanning 20 years,
precise labels,
rich metadata, 
long-tailed distribution, 
distribution shifts over time, 
the visual similarity between the categories,
and multimodal nature, it is a unique addition to the existing benchmarks.

In the provided baseline experiments, we demonstrate how challenging the FungiTastic Benchmarks are. Even state-of-the-art architectures and methods yield modest F-scores of 39.8\% in closed-set classification and 9.1\% in few-shot learning, highlighting the dataset’s challenging nature compared to traditional benchmarks such as CUB-200-2011, Stanford Cars, and FGVC Aircraft.
The proposed zero-shot baseline for the simplest segmentation task, binary segmentation of fungi fruiting body, achieved an average IoU of 89.36\%, which still shows the potential for improvement in fine-grained visual segmentation of fungi.
The open-set baselines show that discovering novel classes remains a difficult task, demanding new techniques tailored to fine-grained recognition.
Additionally, results with non-domain-specific vision-language models reveal a surprisingly strong performance of such models. The fusion experiments of VLMs with supervised models confirm the challenge of accurately classifying rare species in highly imbalanced datasets.

\textbf{Limitations} lie in the data collection process, which affects the overall distribution. Most of the data comes from Denmark, and bias is further introduced through "\textit{random}" sampling. Therefore, some species are more common in frequently sampled areas or are favored by collectors. Some recent observations also miss metadata, which can reduce the effectiveness of classification methods that rely on it.

\textbf{Future work} includes setting up and running future challenges \cite{lifeclef25teaser}, expanding baseline models, adding new test sets, and exploring extra data like traits and species descriptions to enhance multi-modal performance.

\section*{Acknowledgement} 
{\small
This research was supported by the Technology Agency of the Czech Republic, project No. SS73020004.
We extend our sincere gratitude to the mycologists from the Danish Mycological Society, particularly Jacob Heilmann-Clausen, Thomas Læssøe, Thomas Stjernegaard Jeppesen, Tobias Guldberg Frøslev, Ulrik Søchting, and Jens Henrik Petersen, for their contributions and expertise. We also thank the dedicated citizen scientists whose data and efforts have been instrumental to this project. Your support and collaboration have greatly enriched our work and made this research possible. Thank you for your commitment to advancing ecological understanding and conservation.
}
{
    \small
    \bibliographystyle{ieeenat_fullname}
    \bibliography{main}
}


\appendix
\newpage
\section{Evaluation Metrics}
\label{app:eval_metrics}

The diversity and unique features of the FungiTastic dataset allow for the evaluation of various fundamental computer vision and machine learning problems. We present several distinct benchmarks, each with its own evaluation protocol. This section provides a detailed description of all evaluation metrics for each benchmark.

\subsection{Closed set classification}

For closed-set classification, the main evaluation metric is $\text{F}_{1}^{m}$, i.e., the macro-averaged F$_1$-score, defined as

\begin{equation}
\label{eq:fscore}
\text{F}_{1}^{m} = 
\frac{1}{{C}} \sum_{c=1}^{C} {F}_{c},
\quad 
{F}_{c} =  
\frac{2 {P}_c \cdot {R}_c}
{{P}_c + {R}_c},
\end{equation}
where ${P}_c$ and ${R}_c$ are the recall and precision of class $c$ and ${C}$ is the total number of classes.
Additional metrics of interest are Recall@k, defined as 
\begin{equation}
\label{eq:acc}
\text{Recall}@k = \frac{1}{{N}} \sum_{i=1}^{N} \mathbf{1} \left(y_i \in q_{k}(x_i)\right),
\end{equation}
where ${N}$ is the total number of samples in the dataset, $x_i, y_i$ are the $i$-th sample and its label and $q_{k}(x)$ are the top $k$ predictions for sample $x$.

\subsection{Few-shot classification}
The few-shot classification challenge does not have any unknown classes and can be considered as closed-set classification. Unlike other FungiTastic subsets, the few-shot subset does not suffer from high class imbalance and we choose the Top1 accuracy as the main metric. 
F$_1$ score and Top3 accuracy are also reported. All metrics are as defined in closed-set classification.

\subsection{Open-set classification}

The primary metric used for evaluation is the Receiver Operating Characteristic Area Under the Curve (ROCAUC), which measures the ability of the model to distinguish between classes across various threshold values. ROCAUC is defined as the area under the ROC curve, which plots the True Positive Rate (TPR) against the False Positive Rate (FPR) at different classification thresholds where
\begin{equation}
\small
\text{TPR} = \frac{\text{True Positives (TP)}}{\text{True Positives (TP)} + \text{False Negatives (FN)}}, 
\end{equation}
\begin{equation}
\small
\text{FPR} = \frac{\text{False Positives (FP)}}{\text{False Positives (FP)} + \text{True Negatives (TN)}}.
\end{equation}

In addition to ROCAUC, the True Negative Rate (TNR) at 95\% TPR (TNR$^{95}$) is also reported. The TNR, also known as specificity, is defined as:

\begin{equation}
\small
\text{TNR} = \frac{\text{True Negatives (TN)}}{\text{True Negatives (TN)} + \text{False Positives (FP)}}.
\end{equation}

The TNR$^{95}$ metric indicates the specificity achieved when the True Positive Rate (TPR) is fixed at 95\%, reflecting the model's ability to minimize false positives while maintaining a high sensitivity.

The F1-score of the unknown-class, ${F}^{u}$, and the F-score over the known classes, ${F}_k$, are also of particular interest, with ${F}_k$ defined as
\begin{equation}
\label{eq:fscore_unk}
{F}_{K} = \frac{1}{{|K|}} \sum_{c \in {K}} 
F_c,
\end{equation}
where ${K}$ = $\{1 \dots C\} \setminus \{u\}$ is the set of known classes. 

\subsection{Classification beyond 0-1 loss function}
\label{eval:cost}

For the classification beyond 0-1 cost, we follow the definition we set for the annual FungiCLEF competition.
A metric of the following general form should be minimized.

\begin{equation}
   \mathcal{L} =  \frac{1}{{N}} \sum_{i=1}^{{N}} {W}(y_i, q_1(x_i)),
\end{equation}
where $N$ is the total number of samples, $(x_i, y_i)$ are the $i$-th sample and its label, $q_1(x)$ is the top prediction for sample $x$ and $W \in \mathbb{R}^{C \times C}$ is the cost matrix, $C$ being the total number of classes.
For the poisonous/edible species scenario, we define the cost matrix as

\begin{equation}
\small
    \text{W}^{p/e} (y, q_1(x)) = 
    \begin{cases}
   0        & \text{if } d(y) = d(q_1(x)) \\
   c_p        & \text{if } d(y) = 1 \text{ and } d(q_1(x)) = 0 ,\\
   c_e        & \text{otherwise} 
  \end{cases}
\end{equation}
where $d(y), y \in \text{C}$ is a binary function that indicates dangerous (poisonous) species ($d(y) = 1$), $c_p = 100$ and $c_e = 1$.

\subsection{Segmentation}

Provided segmentation masks allow the evaluation of many different segmentation scenarios; here, we highlight two. \\

\noindent\textbf{Binary segmentation}, where the positive class is the foreground (mushroom) and the negative class is the background (the complement). The metric is the intersection-over-union (IoU) averaged over all images ($\text{IoU}_{\text{B}}$), giving each image the same weight

\begin{equation}
    \text{IoU}_{\text{B}} = \frac{1}{I} \sum_{i=1}^{I} \frac{|P_i \cap G_i|}{|P_i \cup G_i|},
\end{equation}
where
 $I$  is the total number of images, 
  $P_i$ is the predicted set of foreground pixels for image  $i$,
  $G_i$ is the ground truth set of foreground pixels for image $i$,
  $|P_i \cap G_i|$ is the intersection (true positives) for image $ i$ and
  $|P_i \cup G_i|$ is the union (true positives + false positives + false negatives) for image $i$. \\

\noindent For \textbf{semantic segmentation}, we adopt the standard mean intersection-over-union (mIoU) metric, where per-class IoUs are averaged, giving each class the same weight
\begin{equation}
    \text{mIoU} = \frac{1}{C} \sum_{c=1}^{C} \frac{|P_c \cap G_c|}{|P_c \cup G_c|},
\end{equation}
where
 $C$  is the total number of classes, 
 $P_c$ is the predicted set of pixels for class $c$, 
 $G_c$ is the ground truth set of pixels for class $c$, 
 $|P_c \cap G_c|$  is the intersection (TPs) and 
 $|P_c \cup G_c|$ is the union (TPs + FPs + FNs).

\section{Supporting Figures and Tables}

\subsection{Closed-set experiment with higher input size}
Following the results provided in the paper, we further experimented with how input size affects classification performance. Switching from 224$\times$224 to 384$\times$384 increased the performance by around five percentage points in all measured metrics and for almost all the architectures. Still, the best-performing model, i.e., BEiT-Base/p16 achieves "\textit{just}" 75\% accuracy and less then 50\% in terms of  $\text{F}_{1}^{m}$.

\begin{table}[h]
\footnotesize
\begin{center}
\setlength{\tabcolsep}{0.6em} %
\renewcommand{\arraystretch}{1.17}
\caption{
\textbf{Closed-set fine-grained classification FungiTastic and FungiTastic--M}. A set of selected state-of-the-art CNN- and Transformer-based architectures evaluated on the test sets. All reported metrics show the challenging nature of the dataset.
}
\label{tab:results_closed_set2}
\begin{tabular}{@{}l| c  c  c | c  c  c@{}}
\toprule
  & \multicolumn{3}{c|}{\scriptsize FungiTastic--M -- 384$^2$} & \multicolumn{3}{c@{}}{\scriptsize FungiTastic -- 384$^2$}  \\  
    \multicolumn{1}{@{}l|}{\textit{Architectures}} & \textbf{Top1} & \textbf{Top3} & \,\,\textbf{$\text{F}_{1}^{m}$}\,\, & \textbf{Top1} & \textbf{Top3} & \,\,\textbf{$\text{F}_{1}^{m}$}\,\, \\  
	\midrule
        ResNet-50           & 66.3 & 82.9 & 39.8 & 66.9 & 80.9 & 36.3 \\
	ResNeXt-50          & 67.0 & 84.0 & 39.9 & 68.1 & 81.9 & 37.5 \\
	EfficientNet-B3     & 67.4 & 82.8 & 40.5 & 68.2 & 81.9 & 37.2 \\
	EfficientNet-v2-B3~~~~  & 70.3 & 85.8 & 43.9 & 72.0 & 84.7 & 41.0 \\
        ConvNeXt-Base       & 70.2 & 85.7 & 43.9 & 70.7 & 83.8 & 39.6 \\
        \midrule
        ViT-Base/p16        & \underline{73.9} & \underline{87.8} & 46.3 & \underline{74.9} & \underline{86.3} & \underline{43.9} \\
        Swin-Base/p4w12     & 72.9 & 87.0 & \underline{47.1} & 74.3 & \underline{86.3} & 43.3 \\
 	BEiT-Base/p16       & \textbf{74.8} & \textbf{88.3} & \textbf{48.5} & \textbf{75.3} & \textbf{86.7} & \textbf{44.5} \\
   \bottomrule
\end{tabular}
\end{center}
\vspace{-0.25cm}
\end{table}

\subsection{FungiTastic -- Dataset statistics}
The FungiTastic dataset offers a rich and diverse collection of observations and metadata. To provide a clearer understanding of its scope, Table \ref{tab:FungiTastic_subsets} presents a statistical overview of its subsets, including the number of observations, associated images, species categories, and metadata availability. Each subset caters to specific benchmarking needs, ensuring comprehensive evaluation scenarios. 

\begin{table}[!h]
\footnotesize
\caption{\textbf{\acs{fungi} dataset splits -- statistical overview.} The number of observations, images, and classes for each benchmark and the corresponding dataset. "\textbf{Unkn}own classes" are those with no available data in training. DNA stands for DNA-sequenced data.}
\setlength{\tabcolsep}{3pt}
    \centering
\begin{tabular}{@{}llrrrr|c@{\hspace{1mm}}c@{\hspace{1mm}}c@{}}
\toprule

 & & &  &  & &  &  & \\ 
\textbf{Dataset} & \textbf{Subset} & \textbf{Observ.} & \textbf{Images} & \textbf{Classes} & \textbf{Unkn.} & \rotatebox[origin=l]{90}{\hspace{-2mm} Metadata} & \rotatebox[origin=l]{90}{\hspace{-2mm} Masks} & \rotatebox[origin=l]{90}{\hspace{-2mm} Captions} \\ 
\midrule
\multirow[c]{3}{*}{\acs{fungi}}    & Train.   & 246,884 & 433,701 & 2,829 & ---  & \checkmark & -- & \checkmark  \\
\multirow[c]{3}{*}{\textit{Closed Set}}    & Val.      &  45,613 &  89,659 & 2,306 & --- & \checkmark & -- & \checkmark   \\
                                     & Test     &  48,378 &  91,832 & 2,336 & --- & \checkmark & -- & \checkmark   \\
                                     & Test$^{\text{DNA}}$    &   2,041 & 5,105 & 725 & --- & \checkmark & -- & \checkmark \\
                         
\midrule
\multirow[c]{3}{*}{\acs{fungim}}    & Train.    & 25,786 & 46,842 & 215 & --- & \checkmark & \checkmark & \checkmark  \\
\multirow[c]{3}{*}{\textit{Closed Set}}    & Val.      &  4,687 &  9,412 & 193 & --- & \checkmark & \checkmark & \checkmark   \\
                                       & Test     &  5,531 & 10,738 & 196 & --- & \checkmark & \checkmark & \checkmark   \\
                                       & Test$^{\text{DNA}}$    &    211 &    642 & 93 & --- & \checkmark & \checkmark & \checkmark \\

\midrule
\multirow[c]{2}{*}{\acs{fungif}} & Train. & 4,293 & 7,819 & 2,427 & --- & \checkmark & -- & \checkmark  \\
\multirow[c]{2}{*}{\textit{Closed Set}}   & Val.   & 1,099 & 2,285 &   570 & --- & \checkmark & -- & \checkmark   \\
                                        & Test  &   999 & 1,911 &   567 & --- & \checkmark & -- & \checkmark  \\
\midrule
\multirow[c]{2}{*}{\acs{fungi}} & Train. & 246,884 & 433,702 & 2,829 & --- & \checkmark & -- & \checkmark   \\
\multirow[c]{2}{*}{\textit{Open Set}} & Val.   &  47,450 &  96,756 & 3,360 & 1,053 & \checkmark & -- & \checkmark   \\
                                     & Test  &  50,084 &  97,551 & 3,349 & 1,000 & \checkmark & -- & \checkmark   \\
\midrule
\multicolumn{2}{@{}l}{\textbf{Total unique values:}} & \textbf{349,307} & \textbf{632,313} & \textbf{6,034} & \multicolumn{1}{r}{\textbf{1,678}} &  &  &    \\
\bottomrule
\end{tabular}
    \label{tab:FungiTastic_subsets}
\end{table}

\subsection{Additional figures}
\begin{figure}[h]
    \centering
    \includegraphics[width=0.95\linewidth]{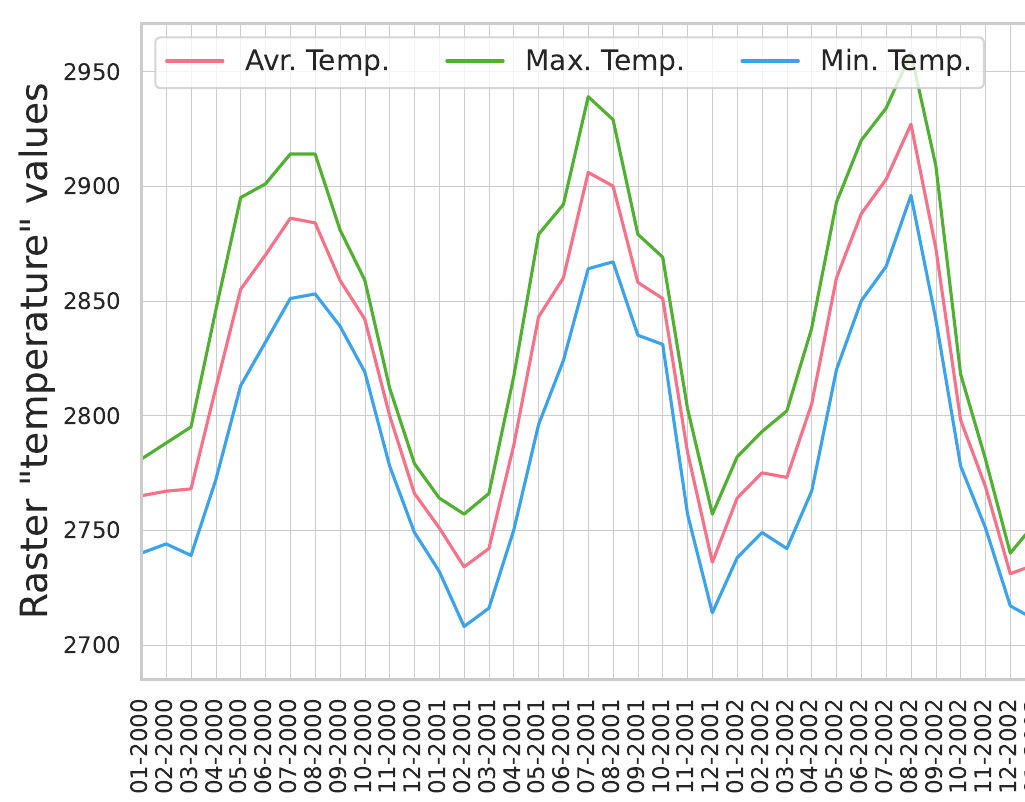}
    \caption{\textbf{Sample of available temperature data}. First 3 years (out of 20 available years) of average, maximum, and minimum monthly temperatures for a randomly selected location.}
    \label{fig:temperature_graph}
\end{figure}

\begin{figure*}[!h]
\begin{tabular}{@{\hspace{1mm}}p{3cm}@{\hspace{3mm}}p{13.5cm}@{}}
\small
\includegraphics[width=3cm]{figures/fig-fusion/0-3122834346.JPG} & \vspace{-3cm}
        \textit{The image shows two mushrooms growing on a dark, decaying log. The mushroom on the left is larger and more prominent. It has a light brown cap with a slightly darker brown rim. The cap's surface appears smooth and slightly shiny. Its stem is thick and light brown, with a hint of green at the base. The smaller mushroom on the right has a similar light brown cap, but its rim is more pronounced and has a white, almost translucent appearance. This gives it a delicate, lacy look. The stem of this mushroom is thinner and lighter in color compared to its larger companion. Both mushrooms have a gilled structure visible underneath their caps. The gills are light brown ......} \\

\includegraphics[width=3cm]{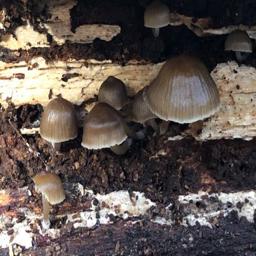} & \vspace{-3cm}
        \textit{The image shows a cluster of mushrooms growing on a decaying log. The fungi have distinctive features that are clearly visible: The caps of the mushrooms are dark brown in color, with a slightly lighter shade around the edges. They have a dome-like shape, resembling small umbrellas or bell-shaped structures. The gills of the mushrooms are visible underneath the caps. They appear as thin, white, vertical lines radiating from the stem to the edge of the cap. These gills are a key identifying feature of many mushroom species. The stems of the mushrooms are light brown in color and appear to be relatively thick and sturdy. They support the caps and extend into ......} \\

\includegraphics[width=3cm]{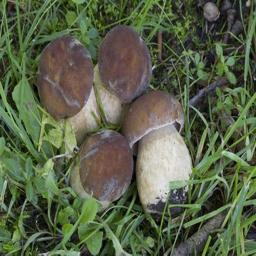} & \vspace{-3cm}
        \textit{The image shows four mushrooms arranged in a square formation on a grassy surface. These fungi have distinctive visual features that are clearly visible: The caps of the mushrooms are dark brown in color, with a smooth, slightly shiny appearance. They have a rounded, dome-like shape that's typical of many mushroom species. The stems of the mushrooms are white, providing a stark contrast to the dark caps. They appear to be relatively thick and sturdy, supporting the caps firmly. The texture of the caps seems to be relatively smooth, without any visible ridges or bumps. The stems appear to have a slightly fuzzy or velvety texture. In terms of size, the mushrooms are of  ......} \\

\includegraphics[width=3cm]{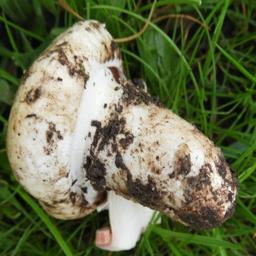} & \vspace{-3cm}
        \textit{The image shows a single mushroom lying on its side in a grassy area. The mushroom has a distinctive shape with a white cap and a long, cylindrical stem. The cap is rounded and slightly curved, while the stem is straight and elongated. The mushroom's surface appears smooth and uniform in color, with a clean white appearance. The cap and stem are of similar width, creating a balanced look. The stem extends from the base of the cap and continues downward, with the tip pointing towards the bottom right corner of the image. The mushroom is covered in dirt, particularly on its stem and parts of the cap. This gives it a freshly harvested appearance and ......} \\ 

\includegraphics[width=3cm]{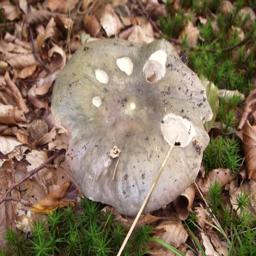} & \vspace{-3cm}
        \textit{The image shows a large, flat mushroom as the central focus. Its cap is predominantly gray with a slightly mottled appearance, featuring some darker patches. The surface of the cap appears smooth and slightly shiny, reflecting light in certain areas. The mushroom's cap is circular in shape, though it's not perfectly round. It has a slightly irregular edge, giving it a natural, organic look. The cap's size is quite substantial, dominating the frame of the image. On the top of the mushroom cap, there are several small, white, circular structures. These are likely the gills or spore-bearing areas of the mushroom. They stand out in contrast against the gray cap, ......} \\
        
\includegraphics[width=3cm]{figures/fig-fusion/0-3122834346.JPG} & \vspace{-3cm}
        \textit{The image shows two mushrooms growing on a dark, decaying log. The mushroom on the left is larger and more prominent. It has a light brown cap with a slightly darker brown rim. The cap's surface appears smooth and slightly shiny. Its stem is thick and light brown, with a hint of green at the base. The smaller mushroom on the right has a similar light brown cap, but its rim is more pronounced and has a white, almost translucent appearance. This gives it a delicate, lacy look. The stem of this mushroom is thinner and lighter in color compared to its larger companion. Both mushrooms have a gilled structure visible underneath their caps. The gills are light brown ......}

\end{tabular}
\vspace{-0.2cm}
\caption{\textbf{Additional image caption samples.} For each photograph, we provide a Malmo-7B image caption-like text description.
}
\label{fig:image-captioning2}
\vspace{-0.5cm}
\end{figure*}

\begin{figure*}[h]
    \centering
    \includegraphics[height=4.0cm, width=4.4cm]{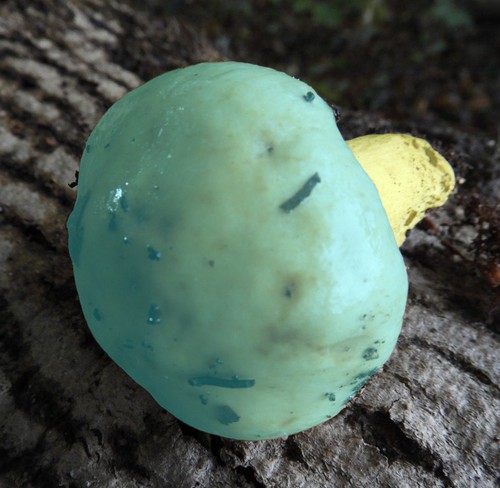}
    \includegraphics[height=4.0cm]{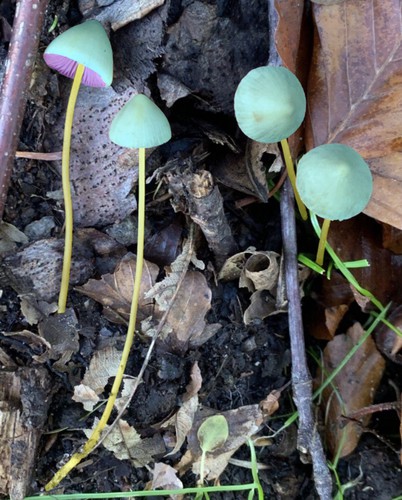}
    \includegraphics[height=4.0cm]{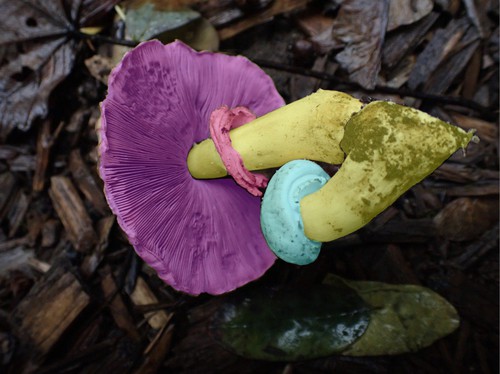} \\
    \vspace{1pt}
    \includegraphics[height=4.0cm, width=2.0cm]{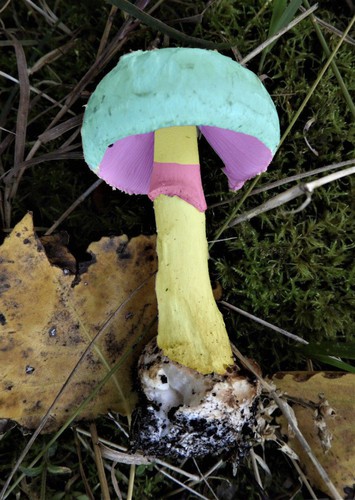} 
    \includegraphics[height=4.0cm]{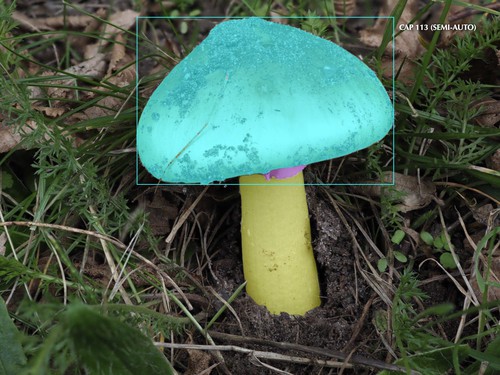}
    \includegraphics[height=4.0cm]{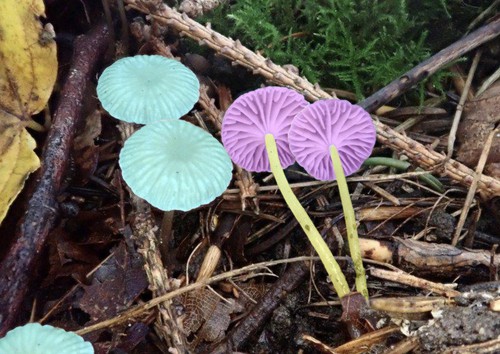}  \\
    \vspace{1pt}
    \includegraphics[height=4.0cm]{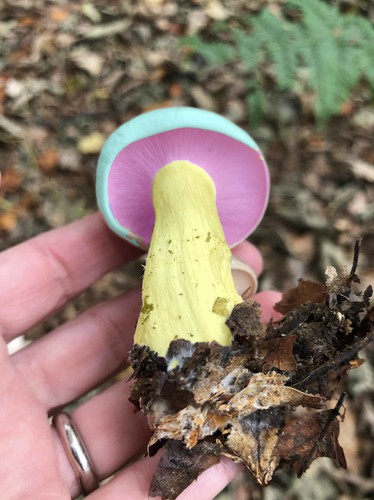} 
    \includegraphics[height=4.0cm, width=5.0cm]{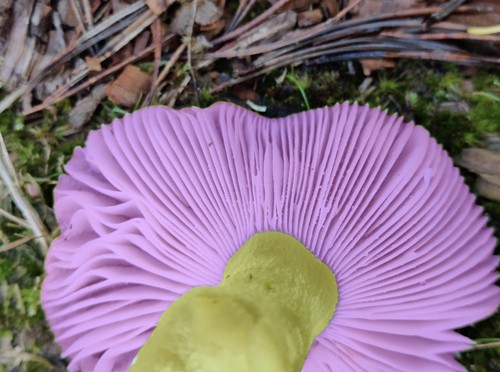} 
    \includegraphics[height=4.0cm, width=5.0cm]{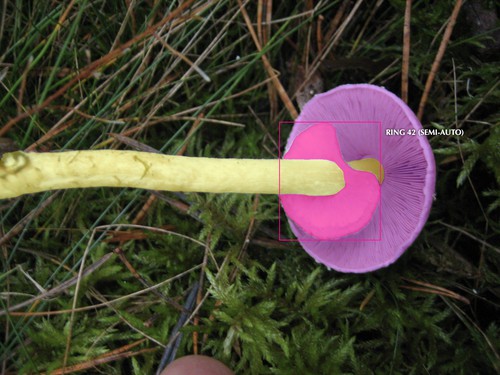} \\
    \vspace{1pt}
    \includegraphics[height=4.0cm]{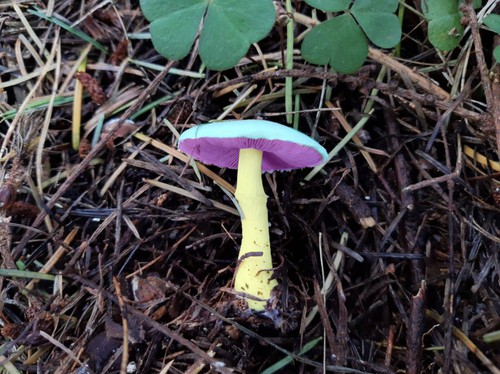} 
    \includegraphics[height=4.0cm]{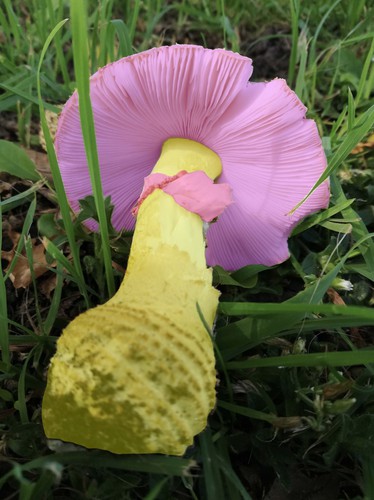} 
    \includegraphics[height=4.0cm, width=4.65cm]{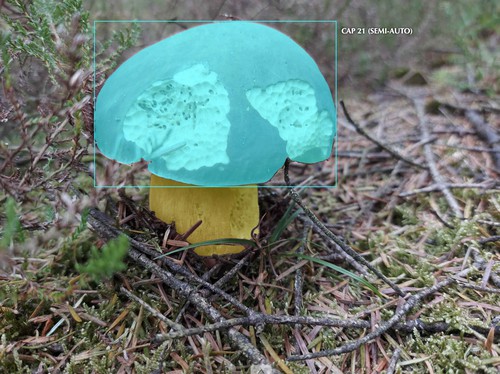} \\
    \vspace{1pt}
    \includegraphics[height=4.0cm, width=5.1cm]{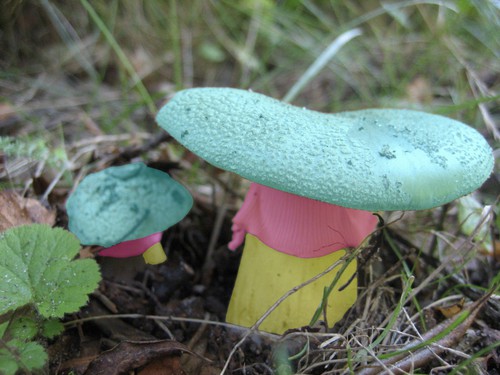} 
    \includegraphics[height=4.0cm]{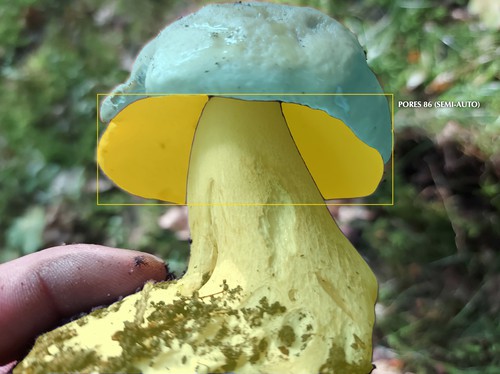} 
    \includegraphics[height=4.0cm]{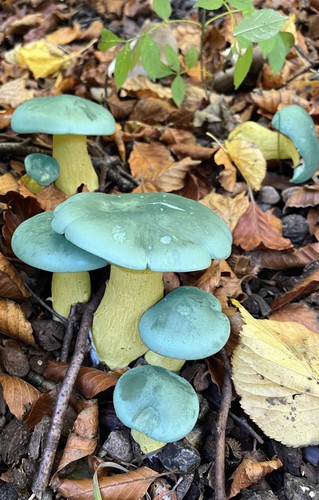} \\
    \vspace{1pt}
    \caption{\textbf{Additional samples of ground truth fruiting body part segmentation}.}
    \label{fig:masks_images}
    \vspace{-0.25cm}
\end{figure*}

To further highlight the unique features of the FungiTastic dataset, we provide additional figures. These include: (i)\,a\,time series sample of temperature data illustrating climatic variability over a 3-year period (Figure \ref{fig:temperature_graph}), (ii) examples of detailed text descriptions generated for individual images to aid in species identification (Figure \ref{fig:image-captioning2}), and (iii) visual samples of ground truth segmentations that highlight different fruiting body parts of fungi (Figure \ref{fig:masks_images}).

\end{document}